%% file: _main.tex
\input{_constants}
\arxiv
\pdfoutput=1
\documentclass[10pt,twocolumn,letterpaper]{article}
\input{cvpr_header}
\begin{document}
%% TITLE
\title{\paperTitle}
\author{\authorBlock}
\maketitle

\input{00_abstract}

\input{01_intro}

\input{02_related}

\input{03_method}

\input{04_experiments}

\input{10_conclusion}

\newpage
\clearpage

{\small
\bibliographystyle{ieeenat_fullname}
\bibliography{11_references}
}

\ifarxiv \clearpage \appendix \input{12_appendix} \fi

\end{document}

%% file: _constants.tex
\def\paperID{8275}
\def\confName{CVPR}
\def\confYear{2024}

\def\paperTitle{Mitigating Object Hallucinations in Large Vision-Language Models through Visual Contrastive Decoding}

\def\authorBlock{
    Sicong Leng$^{1,2,}$\thanks{Equal contribution. Sicong Leng is under the joint PhD program between Alibaba and NTU.} \qquad
    Hang Zhang$^{1,3,}$\footnotemark[1] \qquad
    Guanzheng Chen$^{1,3}$ \qquad
    Xin Li$^{1,3,}$\thanks{Correspondence: \texttt{xinting.lx@alibaba-inc.com}.} \\
    Shijian Lu$^2$ \qquad
    Chunyan Miao$^2$ \qquad
    Lidong Bing$^{1,3}$ \\
    $^1$DAMO Academy, Alibaba Group \qquad $^2$Nanyang Technological University \\
    $^3$Hupan Lab, 310023, Hangzhou, China \\ \\
    {\tt \url{https://github.com/DAMO-NLP-SG/VCD}}
}

\newif\ifreview 
\newif\ifarxiv \newcommand{\arxiv}{\arxivtrue}
\newif\ifcamera 
\newif\ifrebuttal 

%% file: cvpr_header.tex
\ifreview \usepackage[review]{cvpr} \fi
\ifarxiv \usepackage[pagenumbers]{cvpr} \fi
\ifrebuttal \usepackage[rebuttal]{cvpr} \fi
\ifcamera \usepackage{cvpr} \fi

\input{_macros}  % Add packages to _macros.tex

\usepackage{xr-hyper}

\makeatletter
\newcommand*{\addFileDependency}[1]{
  \typeout{(#1)}
  \@addtofilelist{#1}
  \IfFileExists{#1}{}{\typeout{No file #1.}}
}

\makeatother

\definecolor{cvprblue}{rgb}{0.21,0.49,0.74}
\usepackage[pagebackref,breaklinks,colorlinks,citecolor=cvprblue]{hyperref}
\usepackage[listings]{tcolorbox}
\usepackage[capitalize]{cleveref}
\crefname{section}{Sec.}{Secs.}
\crefname{table}{Table}{Tables}
\crefname{figure}{Fig.}{Figs.}

%% file: _macros.tex
\usepackage{graphicx}	
\usepackage{amsmath}	
\usepackage{amssymb}	
\usepackage{booktabs}
\usepackage{times}
\usepackage{microtype}
\usepackage{epsfig}
\usepackage[table,xcdraw,dvipsnames]{xcolor}
\usepackage{caption}
\usepackage{float}
\usepackage{placeins}
\usepackage{color, colortbl}
\usepackage{stfloats}
\usepackage{enumitem}
\usepackage{tabularx}
\usepackage{xstring}
\usepackage{multirow}
\usepackage{xspace}
\usepackage{url}
\usepackage{subcaption}
\usepackage{xcolor}
\usepackage[hang,flushmargin]{footmisc}

\usepackage{multirow}
\usepackage{booktabs}

\ifcamera \usepackage[accsupp]{axessibility} \fi

% ...

\ifarxiv  \fi

\newcommand{\R}[1]{{%
    \textbf{%
        \ifstrequal{#1}{1}{\textcolor{red}{R#1}}{%
        \ifstrequal{#1}{2}{\textcolor{blue}{R#1}}{%
        \ifstrequal{#1}{3}{\textcolor{magenta}{R#1}}{%
        \ifstrequal{#1}{4}{\textcolor{teal}{R#1}}{%
                           \textcolor{cyan}{R#1}%
        }}}}%
    }%
}}

%% file: 00_abstract.tex
\begin{abstract}
% Abstract goes here.
%(\textbf{brief intro of hallucination})
%\textbf{(no other modules needed and orthogonal to revisor solution)}
Large Vision-Language Models (LVLMs) have advanced considerably, intertwining visual recognition and language understanding to generate content that is not only coherent but also contextually attuned. Despite their success, LVLMs still suffer from the issue of object hallucinations, where models generate plausible yet incorrect outputs that include objects that do not exist in the images. 
To mitigate this issue, we introduce Visual Contrastive Decoding (VCD), a simple and training-free method that contrasts output distributions derived from original and distorted visual inputs.
The proposed VCD effectively reduces the over-reliance on statistical bias and unimodal priors, two essential causes of object hallucinations.
This adjustment ensures the generated content is closely grounded to visual inputs, resulting in contextually accurate outputs.
Our experiments show that VCD, without either additional training or the usage of external tools, significantly mitigates the object hallucination issue across different LVLM families. 
Beyond mitigating object hallucinations, VCD also excels in general LVLM benchmarks, highlighting its wide-ranging applicability.
Codes will be released.

\end{abstract}

%% file: 01_intro.tex
\section{Introduction}
\label{sec:intro}
%(\textbf{Opening Statement})
%\lx{Note: should we use LVLMs (instead of VLMs) across the paper? Just to be consistent}
Large Vision-Language Models (LVLMs) have become integral in the intersection of computer vision and natural language processing, enabling a range of applications due to their ability to generate contextually relevant textual descriptions from visual inputs. These models are characterized by their effectiveness in capturing and translating complex visual patterns into coherent linguistic representations~\cite{liu2023visual,zhu2023minigpt4,ye2023mplugowl,li2023otter,dai2023instructblip,gong2023multimodalgpt,maaz2023videochatgpt,zhang2023videollama,bai2023qwen}.
%VLMs have proven essential in a variety of application domains, including but not limited to, content creation, image and video annotation, and interactive systems where understanding and interpreting visual content is crucial.
The evolution of LVLMs is marked by ongoing improvements in model architecture, training methodologies, and data diversity, leading to enhanced performance and application versatility. Despite these advancements, specific challenges persist, with the issue of object hallucination~\cite{li2023evaluating,gunjal2023detecting,liu2023mitigating,lovenia2023negative} being a prominent concern that impacts the reliability and applicability of LVLMs across domains.

\input{figs/figure1}

%(\textbf{Problem Statement})
Object Hallucination in this context refers to the phenomenon where LVLMs generate textual content that is semantically coherent but inconsistent with ground-truth objects in the given image.
%\lx{$\Rightarrow$ where VLMs generate textual contents that semantically coherent but inconsistent with the ground-truth object in the given image}. 
%This phenomenon manifests as generating plausible descriptions or answers that do not accurately represent the visual data. 
%(\textbf{TODO: Existing works proving the severe problem of hallucinations in VLMs?})
This challenge not only reveals fundamental issues of LVLMs, such as over-reliance on statistical bias~\cite{agarwal2020towards,agrawal2016analyzing,goyal2017making,li2023evaluating} and unimodal priors~\cite{yan2023overcoming,zhibo2023overcoming,han2022visual,wu2022overcoming,gupta2022swapmix,niu2021counterfactual}, but also has direct implications for the practical deployment of LVLMs. In applications where precision and reliability of generated content are paramount, object hallucinations can lead to misinformation, misinterpretation, and subsequent erroneous decision-making. In domains like healthcare~\cite{wang2023chatcad,hu2023advancing}, autonomous systems~\cite{chen2023driving,wu2023embodied}, and robotics~\cite{mai2023llm,liu2023llm}, such inaccuracies are not just undesirable but could have significant consequences.
Addressing the hallucination issue is therefore essential to enhance the integrity, reliability, and broad applicability of LVLMs in various real-world scenarios.
%\lx{Note: ``real-world scenarios'' should be enough}.

%(\textbf{Existing Solutions and Their Limitations})
Various approaches have been explored to curb object hallucinations in VLMs. Early works made attempts on small-scale VLMs by either performing fine-grained modality alignment~\cite{biten2022let} or reducing the statistical bias of object co-occurrence with data augmentation~\cite{rohrbach2018object,kim2023exposing}. However, the behaviors of LVLMs differ significantly from small-scale VLMs, making related methods impractical to generalize and scale up~\cite{kaplan2020scaling,wei2022emergent}. Several recent studies address this issue by proposing hallucination-targeted datasets for fine-tuning~\cite{liu2023aligning,gunjal2023detecting}, training a post-hoc revisor to reconstruct less hallucinatory outputs~\cite{zhou2023analyzing} or adapting factually augmented Reinforcement Learning from Human Feedback (RLHF)~\cite{sun2023aligning}. 
%Despite their effectiveness, the data annotation and RLHF process are time-consuming and labor-intensive, requiring human expertise and effort. 
%(\textbf{Objective of the Current Study})
While existing interventions for object hallucination in LVLMs have shown effectiveness, the incurred human effort and computational cost highlight a pressing need for a simpler but efficient approach.
%This paper aims to contribute to this ongoing discourse, enhancing VLMs' performance across various applications. 
%We propose targeted solutions and seek to explore and analyze the underlying causes of hallucination, fostering a comprehensive understanding that is integral for developing effective and robust mitigation techniques.
%  ，we found that the visual uncertainty will amplifier the hallucination。 When the Lvlm received the distorted image input,  他往往会根据经验进行胡说八道（language prior and static bais）。 这也是幻觉产生的根本原因之一。Can we utilize the amplified bais brought from the distorted image
%(\textbf{Introduction and Significance of the Current Study})

In this work, we analyze the effect of visual uncertainty on the two primary causes of object hallucinations in LVLMs, namely statistical bias and unimodal priors (i.e., language priors). 
%Building on these insights
Building on the analysis above, we introduce Visual Contrastive Decoding (VCD), a training-free technique designed to mitigate object hallucination in LVLMs. As shown in Figure~\ref{fig:vcd illustration}, VCD is grounded in the principle of contrasting output distributions from original and distorted visual inputs.
%, focusing on the comparative analysis of models’ responses to original and distorted visual inputs.
Hence, it acts as a corrective mechanism and calibrates the model’s over-reliance on language priors from integrated LLMs and statistical bias of LVLMs' pretraining corpus. 
%By deliberately contrasting the model’s responses to original and distorted visual signals, we instigate a refined alignment between the two input modalities. The model thus becomes adept at weaving visual perceptions seamlessly into linguistic expressions, elevating the authenticity and relevance of its generative capabilities.
In the realm of efficiency, VCD stands out due to its minimal computational overhead compared with previous studies~\cite{liu2023aligning,gunjal2023detecting,zhou2023analyzing,sun2023aligning}, circumventing the need for additional training or the usage of external tools (e.g., other pretrained models). 
Our experiments demonstrate VCD's effectiveness, with consistent improvements on multiple object hallucination benchmarks (e.g., up to $+7.4$ F1 score boost on POPE~\cite{li2023evaluating} and $+18\%$ improvement on MME~\cite{fu2023mme}) across different LVLM families, including LLAVA-1.5~\cite{liu2023visual,liu2023improved}, InstructBLIP~\cite{dai2023instructblip}, and Qwen-VL~\cite{bai2023qwen}. In addition, our method is also beneficial to the general perception capacities of LVLMs as evidenced by benchmarking on MME and LLaVA-Bench\footnote{\url{https://huggingface.co/datasets/liuhaotian/llava-bench-in-the-wild}}, indicating its potential applicability beyond the scope of object hallucination mitigation. 
% Additionally, the flexible nature of VCD allows for an in-depth exploration of the causative factors of object hallucinations. By applying distortions to the visual inputs in different granularities, we gain nuanced insights into the model's response dynamics. This adaptability not only amplifies VCD's utility but also contributes to the broader understanding of object hallucination phenomena in VLMs, laying a foundation for future innovations and refinements in this domain.

%By mitigating the hallucination issue, our research stands to fortify the reliability of VLMs, with profound implications for applications ranging from automated content generation to augmented reality, where accuracy and context are paramount.

%(\textbf{Summary\&Structure of the Paper})
To sum up, our main contributions are as follows:
\begin{enumerate}
    \item We conduct an in-depth analysis of the effect of visual uncertainty on object hallucinations in LVLMs, particularly from the aspects of statistical bias and unimodal priors. 
    \item Inspired by the analysis above, we design VCD, a training-free technique that can effectively mitigate object hallucinations in LVLMs. It calibrates the model's outputs by contrasting output distributions derived from original and distorted visual inputs, ensuring more consistent content generation.
    \item Through comprehensive experiments, we demonstrate the efficacy of the proposed VCD in alleviating object hallucination and enhancing general perception capability. Our method yields notable improvements without the need for additional training or external tools.
    % \item VCD’s flexible design facilitates a deeper investigation into the underlying causes of object hallucinations. Our analysis, supported by varied distortions to visual inputs, offers nuanced insights and broadens the understanding of object hallucination phenomena in VLMs (Section~\ref{sec:discussions}).
\end{enumerate}

% To insert a figure: \input{figs/template}
% Or table: \input{tables/template}

%% file: figs/figure1.tex
\begin{figure}[tp]
    \centering
    \includegraphics[width=1\linewidth]{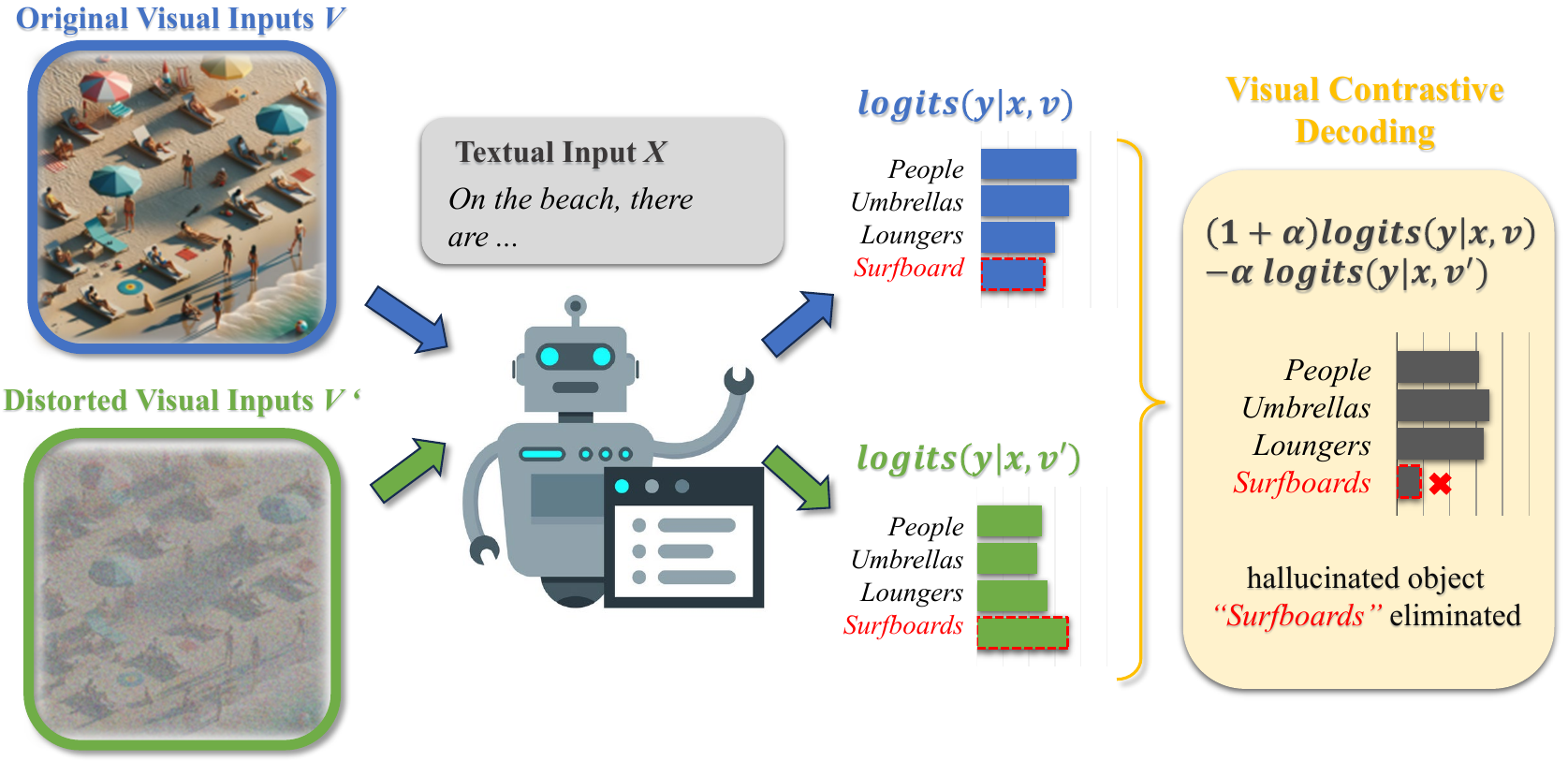}
    \caption{An illustration of Visual Contrastive Decoding. The hallucinated object \textit{``Surfboards"} is highlighted in \color{red}{red}\color{black}, and it is eliminated during the generative process by contrasting with the output distribution that favors hallucinations.}
    \label{fig:vcd illustration}
    \vspace{-0.4cm}
\end{figure}

%% file: 02_related.tex
\section{Related Work}
\label{sec:related}
\subsection{Visual-Language Models}
The development of Vision-Language Models (VLMs) has transitioned from being rooted in BERT-based language decoders \cite{devlin2018bert,liu2019roberta,koroteev2021bert} for merging visual and textual data \cite{li2019visualbert,sun2019videobert,wang2022git,li2022blip}, to a notable advancement ushered by the integration of Large Language Models (LLMs) \cite{gilardi2023chatgpt,touvron2023llama,tay2022ul2,raffel2020exploring,brown2020language,chowdhery2022palm,alpaca,vicuna2023,bai2023qwenllm}. The advent of LLMs heralded the emergence of Large Vision-Language Models (LVLMs) \cite{alayrac2022flamingo,chen2022pali,driess2023palm,li2022blip}, characterized by enhanced capabilities and performance. 
%These models, like LLAVA \cite{liu2023visual}, MiniGPT4 \cite{zhu2023minigpt4}, InstructBLIP \cite{}, and mplug-Owl, are distinguished by their incorporation of LLMs, resulting in improved intermodal interactions and task-specific optimizations. 
In this phase, LVLMs, supported by end-to-end training techniques, demonstrated unified decoding of visual and textual tokens, marking a significant enhancement in their performances and adaptability. Recent developments have seen a focus on Visual Instruction Fine-tuning \cite{liu2023visual}, showcasing adaptability to a variety of vision-language tasks. The methodologies adopted, ranging from integrating cross-modal alignment networks to fine-tuning LLaMA models, underscore a trend of diversification and specificity in the approach \cite{dai2023instructblip,li2023otter,ye2023mplugowl,bai2023qwen}. 
%These models, exemplified by their architectural and algorithmic innovations, highlight the ongoing journey of VLMs towards enhanced integration, performance, and application versatility.

\subsection{Hallucination in VLMs}
Prior to the advent of LLMs, the NLP community has primarily defined ``hallucination" as the generation of nonsensical content or content that deviates from its sources \cite{lee2018hallucinations,zhou2020detecting,lin2021truthfulqa,ji2023survey,zhang2023siren,shi2023replug}. In the realm of VLMs, ``object hallucination" is also well-documented, referring to models producing plausible outputs that include objects that do not match or are missing from images \cite{rohrbach2018object,biten2022let,li2023evaluating}. 
Mitigating object hallucination in VLMs has typically involved strategies such as fine-grained contrastive learning \cite{zeng2021multi}, ROI feature fusion \cite{biten2022let}, and the curtailment of co-occurrence patterns via data augmentation \cite{kim2023exposing}. However, with the distinct training paradigms and model architectures that characterize traditional VLMs and contemporary LVLMs, adapting these strategies to the newer auto-regressive approaches in LVLMs poses significant challenges \cite{kaplan2020scaling,wei2022emergent}. 

Recent efforts have sought to navigate these complexities, with studies delving into the evaluation and detection of object hallucinations within the domain of LVLMs \cite{wang2023evaluation,liu2023aligning,li2023evaluating,lovenia2023negative}.
% For example, CHAIR \cite{rohrbach2018object} is proposed to evaluate the exact match between objects residing in generated and ground-truth image captions. Alternatively, 
For example, POPE \cite{li2023evaluating} converts the hallucination into a binary classification problem to probe the model's awareness of whether a specific object exists in the image. 
Concurrently, there has been a notable push towards the development of refined datasets tailored for fine-tuning existing LVLMs \cite{gunjal2023detecting,li2023m,liu2023aligning}, training a post-hoc revisor to detect and reconstruct less hallucinatory outputs \cite{zhou2023analyzing}, and adapting factually augmented RLHF \cite{sun2023aligning}. Nevertheless, existing approaches that acquire additional datasets, conduct fine-grained tuning on original or newly introduced models, or utilize other off-the-shell pretrained models can be time-consuming, labor-intensive, and computationally costly. Instead, we propose a conceptually different and training-free approach, VCD, that contrasts the output distributions with original and distorted visual inputs to calibrate the model's over-reliance on unimodal priors and statistical bias, without utilizing external models.

%% file: 03_method.tex
\section{Method}
\label{sec:method}

\subsection{Decoding of Vision-Language Models}
We consider an LVLM parametrized by \( \theta \). The model takes as input a textual query \( {x} \) and a visual input \( {v} \), where \( {v} \) provides contextual visual information to assist the model in generating a relevant response \( {y} \) to the textual query. 
%We denote the generated textual response as \( {y} \), which is obtained by querying the model with both the textual and visual inputs. 
The response \( {y} \) is sampled auto-regressively from the probability distribution conditioned on the query \( {x} \) and the visual context \( {v} \). Mathematically, this can be formulated as:
\begin{equation}
\begin{aligned}
y_t & \sim p_\theta\left(y_t \mid {v}, {x}, {y}_{<t}\right), \\
& \propto \exp \operatorname{logit}_\theta\left(y_t \mid {v}, {x}, {y}_{<t}\right),
\end{aligned}
\end{equation}
where \( y_t \) denotes the token at time step \( t \), and \( {y}_{<t} \) represents the sequence of generated tokens up to the time step ($t - 1$). 
In the decoding phase of LVLMs, object hallucinations often emerge when probabilities are erroneously allocated to tokens that do not align with the presented visual input $v$. Previous studies have identified two primary causes of this problem: (1) statistical biases inherent in training data (e.g., prevalent but superficial object correlations)~\cite{agarwal2020towards,agrawal2016analyzing,goyal2017making}, and (2) over-reliance on language priors embedded within the powerful LLMs used as decoders~\cite{li2023evaluating,yan2023overcoming,zhibo2023overcoming,han2022visual}. Our approach to mitigate object hallucinations first amplifies these undesirable behaviors with vague inputs and subsequently contrasts with them in the decoding process.

% Previous works find that object hallucinations primarily arise from two factors:  (1) statistical biases in training data and (2) over-reliance on unimodal priors~\cite{agarwal2020towards,agrawal2016analyzing,goyal2017making,li2023evaluating,yan2023overcoming,zhibo2023overcoming,han2022visual,wu2022overcoming,gupta2022swapmix,niu2021counterfactual}.
% Statistical biases are typically presented as frequent but superficial object correlations, often leading to the generation of contextually misaligned objects in the response.
% This issue is further compounded by an overly dependence on unimodal, especially language, priors ingrained during the integrated LLMs' pretraining phase, compelling the models to favor textual patterns over actual visual inputs. 

\input{figs/figure2}
\subsection{Visual Uncertainty Amplifies Hallucinations}
\label{subsec: observation}
The fidelity of visual input is pivotal for LVLMs to accurately encode visual features and generate outputs faithfully. Yet, the introduction of uncertainty in visual inputs can tilt the equilibrium. 
This section delves into a comprehensive analysis aiming to validate the assumption that increased visual uncertainty can amplify the language priors and statistical biases in LVLMs, thus exacerbating object hallucination.

\vspace{0.2cm}
\noindent\textbf{Introduction of Visual Uncertainty}
In this paper, we propose to adopt the most elementary method---applying a Gaussian noise mask to the original image---to introduce visual uncertainty. This method, although straightforward, provides an initial benchmark to estimate the baseline effects of visual uncertainty on model outputs. Following the forward diffusion process in image generation~\cite{ho2020denoising}, the distorted image is modeled as follows:
\begin{equation}
\label{eq:1}
\begin{aligned}
&q\left({v}_t \mid {v}_{t-1}\right)=\mathcal{N}\left({v}_t ; \sqrt{1-\gamma} {v}_{t-1}, \gamma \mathbf{I}\right) \\
&q\left({v}_{T} \mid {v}_0\right)=\prod_{t=1}^T q\left({v}_t \mid {v}_{t-1}\right),
\end{aligned}
\end{equation}
where $v_0$ denotes the original visual input (i.e., original image) and $\mathbf{I}$ refers to an identity matrix. We incrementally add a small amount of Gaussian noise for $T$ steps, producing a sequence of distorted images $v_1,\dots,v_T$. The original image $v_0$ gradually loses its distinguishable features as step $t$ goes larger, where the amount of noise added in each step is controlled by $\gamma$. Eventually, when $T \rightarrow \infty$, visual uncertainty reaches the maximum and $v_T$ will become indistinguishable from Gaussian noise.

%\footnote{While this study employs a basic Gaussian noise approach, more fine-grained techniques, like object-level blurring, hold the potential for improved outcomes. Exploring such distortions is considered a direction for future research.}

\vspace{0.2cm}
\noindent\textbf{Visual Uncertainty Amplifies Language Priors}
Figure~\ref{fig: language priors} shows that visual uncertainty can compel LVLMs to overlook visual evidence and overly exploit language priors for decision-making. However, this tendency is not entirely unexpected, as LLMs are designed to predict next-word probabilities based on vast textual corpora. 
When confronted with ambiguous visual stimuli, an LVLM might misinterpret these conventional, text-based predictions as a ``safety net''. 
These priors, while generally useful, can introduce biases or assumptions that are inconsistent with the actual visual content, particularly when the visual input lacks clarity. 

\vspace{0.2cm}
% % current LVLMs mainly 由强大的语言模型和 and trained on ，some researcher 
\noindent\textbf{Visual Uncertainty Amplifies Statistical Bias}
The construction of most vision-language pretraining datasets is predominantly based on MSCOCO~\cite{lin2014microsoft}, which inherently suffers from an unbalanced object distribution and biased object correlations. 
Previous works \cite{li2023evaluating,zhou2023analyzing} point out that LVLMs, trained on such data, may inherit those statistical biases to generate descriptions with hallucinated objects.
To further examine the hypothesis that visual uncertainty may amplify statistical biases from pretraining, we designed two targeted experiments to verify (1) if LVLMs hallucinate frequent objects more with distorted visual inputs and (2) if LVLMs are more prone to hallucinate objects that frequently co-occur with ground-truth objects in the image with distorted visual inputs.
Figure~\ref{fig:bias} shows an evident tendency that LVLMs are more prone to hallucinate frequent and co-occurring objects, attributing to the imbalanced object distributions and spurious object correlations inherited from the training data.

\subsection{Visual Contrastive Decoding}

%\subsection{Visual Contrastive Decoding}
%\subsection{Contrasting with Visual Uncertainty}
\subsubsection{Contrasting the Predictions}
% Our observations in previous section reveal that LVLMs exhibit a biased output distribution towards hallucinations when presented with distorted visuals, as compared to original visual inputs.
Our observations in the previous section reveal that visual uncertainty not only amplifies reliance on language priors but also makes LVLMs more likely to be biased by superficial object correlations present in pretraining datasets, leading to more severe hallucinations.
%Our observations in Section~\ref{sec:invest} show that LVLMs, when presented with distorted visuals, have a skewed output distribution leaning towards hallucinations. 
% Could we filter out the hallucinations amplified by the distorted input, such as language priors and statistical biases, and generate text from the remaining faithful behaviors of clean visual input?  To operationalize this goal, we introduce Visual Contrastive Decoding (VCD). VCD is formulated to mitigate the hallucination problem by contrasting model outputs generated from both original and distorted visual inputs. 
In light of this, we introduce Visual Contrastive Decoding (VCD). VCD is formulated to counteract the statistical biases and language priors in LVLMs by contrasting model outputs generated from original and distorted visual inputs. 
%Through this contrasting mechanism, VCD diligently curtails hallucinations, ensuring the outputs resonate more faithfully with the given visual context.
This is achieved without necessitating additional training or external pretrained models, making VCD a cost-effective and efficient solution.

% VCD operates by contrasting the output distributions generated from the original visual inputs and their distorted versions. 
Specifically, given a textual query ${x}$ and a visual input ${v}$, the model generates two distinct output distributions: one conditioned on the original ${v}$ and the other on the distorted visual input ${v'}$, which is derived by applying pre-defined distortions (i.e., Gaussian noise mask) to ${v}$. 
Then, a new contrastive probability distribution is computed by exploiting the differences between the two initially obtained distributions. 
%This process calibrates the model’s responses, serving to interrupt the model’s learned biases and priors.
%fostering a more balanced utilization of both visual and textual data during the generation process. 
The new contrastive distribution $p_{vcd}$ is formulated as:
\begin{equation}
\label{eq:3}
\begin{gathered}
p_{vcd}\left(y \mid v, v', x\right) =\operatorname{softmax}\left[ (1+\alpha) 
\operatorname{logit}_\theta\left(y \mid v, x\right) \right.\\
\left.-\alpha \operatorname{logit}_\theta\left(y \mid v', x\right)\right],
\end{gathered}
\end{equation}
where larger $\alpha$ values indicate a stronger amplification of differences between the two distributions ($\alpha=0$ reduces to regular decoding). 
From the adjusted output distribution $p_{vcd}$, we can apply various sampling strategies, such as nucleus sampling~\cite{holtzman2019curious} and beam search~\cite{freitag2017beam}.

Essentially, VCD serves as a corrective mechanism, reducing hallucinations by contrasting against a distribution predisposed to favoring them. Alternatively,
VCD can also be interpreted as a form of contrastive ensemble that differentiates between the logits of $p_\theta\left(y \mid v, x\right)$ and $p_\theta\left(y \mid v', x\right)$. This method echoes the contrastive objective commonly employed in image generation. For instance, classifier-free diffusion models \cite{ho2022classifier} estimate diffusion noise using $(1+\alpha)\epsilon_\theta(x,c)-\alpha\epsilon_\theta(x)$, where $c$ serves as a controlling factor. 
In the realm of text generation, several studies have also exploited contrastive decoding for more faithful generation~\cite{liu2021dexperts,li2022contrastive,o2023contrastive,shi2023trusting}.
% In the realm of text generation, a comparable approach is introduced by \citet{shi2023trusting} propose context-aware decoding with the same intuition, with a focus on contrasting the full input with and without textual context, prompting LLMs to trust the contextual evidence\footnote{Furthermore, rather than relying on a single model $\theta$ in this study, there exists the alternative to employ different models to adjust the output distribution, serving to suppress unwanted model behaviors or distill the expert model's capabilities~\cite{liu2021dexperts,li2022contrastive,o2023contrastive}. }. 
\input{figs/bias}
\subsubsection{Adaptive Plausibility Constraints}
According to the formation of the contrastive distribution $p_{vcd}$ in Equation~\ref{eq:3}, a challenge may arise as it penalizes the model's entire output behaviors influenced by distorted visual inputs. However, this is not universally correct – the output distributions with distorted visual inputs can still uphold fundamental linguistic standards and common sense reasoning. Indiscriminate penalization could inaccurately punish these valid outputs and promote the generation of implausible tokens. To address this issue, we follow \citet{li2022contrastive} to implement an adaptive plausibility constraint that is contingent upon the confidence level associated with the output distribution with original visual inputs:
%This approach mitigates the risk of over-penalization, especially when the model exudes high confidence in the outputs correlated with the original visual inputs. 
\begin{equation}
\label{eq:5}
\begin{gathered}
\begin{aligned}
& \mathcal{V}_{\text {head }}\left(y_{<t}\right)= 
 \{y_t \in \mathcal{V}:  \\
 &p_{\theta}\left(y_t \mid v,x,y_{<t}\right) \geq \beta \max _w p_{\theta}\left(w \mid v,x,y_{<t}\right)\},
\end{aligned}
\\
\begin{aligned}
p_{vcd}\left(y_t \mid v, v', x\right) = 0, \text{ if } y_t \notin \mathcal{V}_{\text {head }}\left(y_{<t}\right),
\end{aligned}
% \begin{aligned}
% & \log p_{vcd}\left(y \mid v, v', x\right) \\
% & = \begin{cases}\log \frac{\left(1+\alpha\right) \cdot p_\theta\left(y \mid v, x\right) }{\alpha \cdot p_\theta\left(y \mid v', x\right)}, & \text { if } y_i \in \mathcal{V}_{\text {head }}\left(y_{<i}\right), \\
% -\inf , & \text { otherwise. }\end{cases}
% \end{aligned}
\end{gathered}
\end{equation}
where $\mathcal{V}$ is the output vocabulary of LVLMs and $\beta$ is a hyperparameter in $[0,1]$ for controlling the truncation of the next token distribution. Larger $\beta$ indicates more aggressive truncation, keeping only high-probability tokens. 
%This ensures a balanced adjustment, where valid linguistic structures are preserved while minimizing the propagation of biases and the overemphasis on unimodal priors.

Combining the visual contrastive decoding and the adaptive plausibility constraint, we obtain the full formulation:
\begin{equation}
\begin{gathered}
y_t \sim \operatorname{softmax}\left[(1+\alpha) \operatorname{logit}_\theta\left(y_t \mid v, x, y_{<t}\right)\right. \\
\left.-\alpha \operatorname{logit}_\theta\left(y_t \mid v', x, y_{<t}\right)\right],\\
{subject \ to} \ y_t \in \mathcal{V}_{\text {head }}\left(y_{<t}\right)
\end{gathered}
\end{equation}

Incorporating adaptive plausibility constraints refines the contrastive distribution, bolstering confidence in straightforward decisions.
%while mitigating the influence of statistical bias and unimodal priors on complex or ambiguous token generations. 
This ensures that when the model is highly confident in its outputs associated with the original inputs, the candidate pool is streamlined, often retaining a singular token with high probability. Such an approach effectively neutralizes potential adverse effects of VCD, preventing it from inadvertently promoting the generation of implausible tokens and maintaining the integrity of the generated content.

%% file: figs/figure2.tex
\begin{figure}[tp]
    \centering
    \includegraphics[width=1\linewidth]{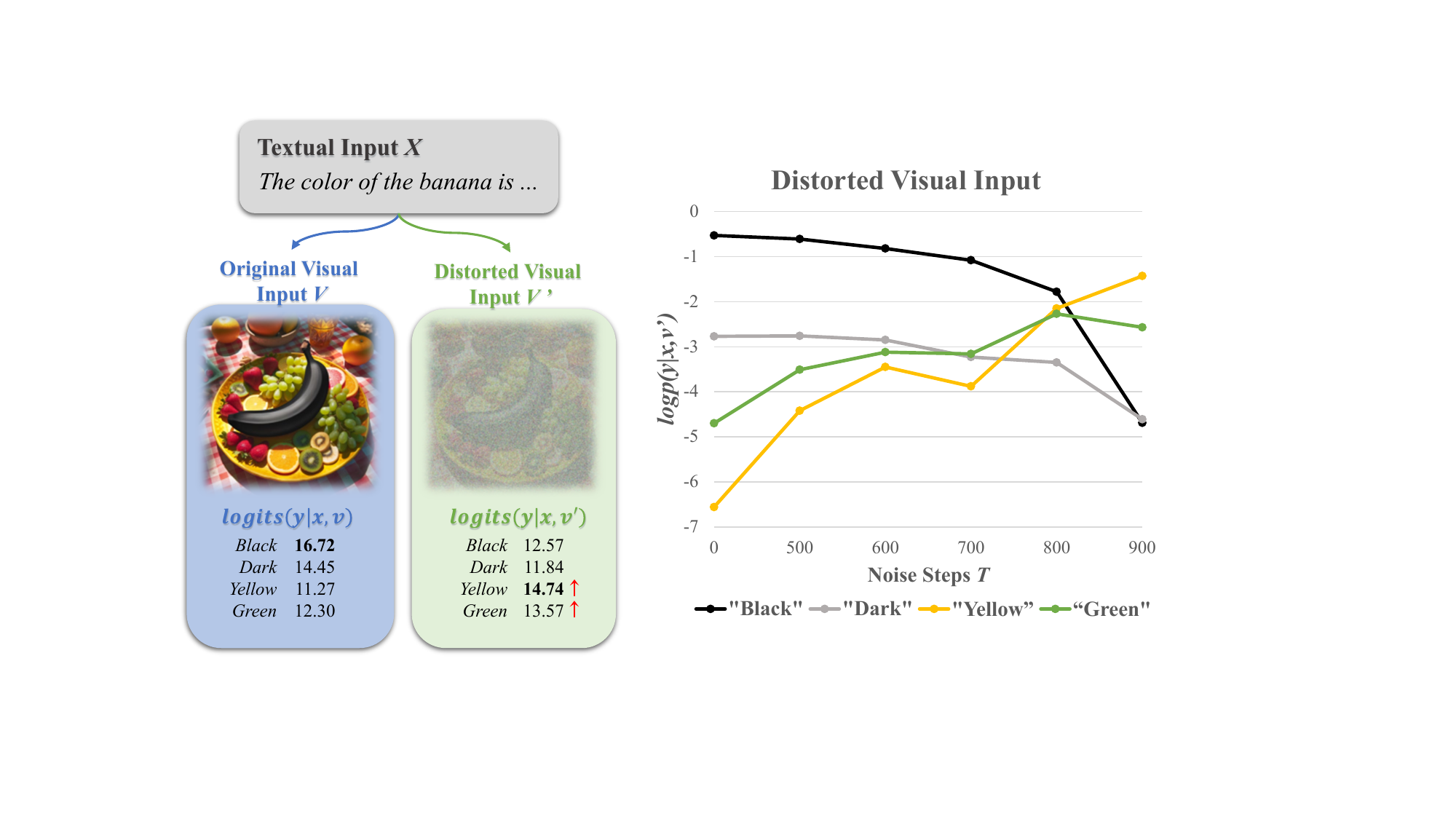}
    \caption{An illustration of visual uncertainty amplifying language priors. Given an image featuring a black banana among other colorful fruits, LVLMs favor more conventional banana colors—such as "\textit{yellow}" and "\textit{green}", with increasing visual uncertainty. The ground-truth color "\textit{black}" diminishes in probability ($logp(y|x,v')$) as the distortion escalates, making LVLMs over-reliant on the language priors from LLM pre-training that typically associate bananas with being yellow or green.}
    \label{fig: language priors}
    \vspace{-0.4cm}
\end{figure}

%% file: figs/bias.tex
% \begin{figure}[h]
%     \centering
%     % First subfigure
%     \begin{subfigure}[b]{0.5\textwidth}
%         \centering
%         \includegraphics[width=\textwidth]{figs/bias1.pdf}
%         \caption{Hallucination times of top three objects.}
%         \label{fig:sub1}
%     \end{subfigure}
%     %\hfill % Optional: add some space between the two subfigures
%     % Second subfigure
%     \begin{subfigure}[b]{0.5\textwidth}
%         \centering
%         \includegraphics[width=\textwidth]{figs/bias2.pdf}
%         \caption{Hallucination times of top three objects co-occurring with "dining table".}
%         \label{fig:sub2}
%     \end{subfigure}

%     % Caption for the whole figure
%     \caption{Hallucination times of frequently appearing/co-occurring objects in MSCOCO.}
%     \label{fig:bias}
%     \vspace{-0.4cm}
% \end{figure}
\begin{figure}[t]
    \centering
    \includegraphics[width=1\linewidth]{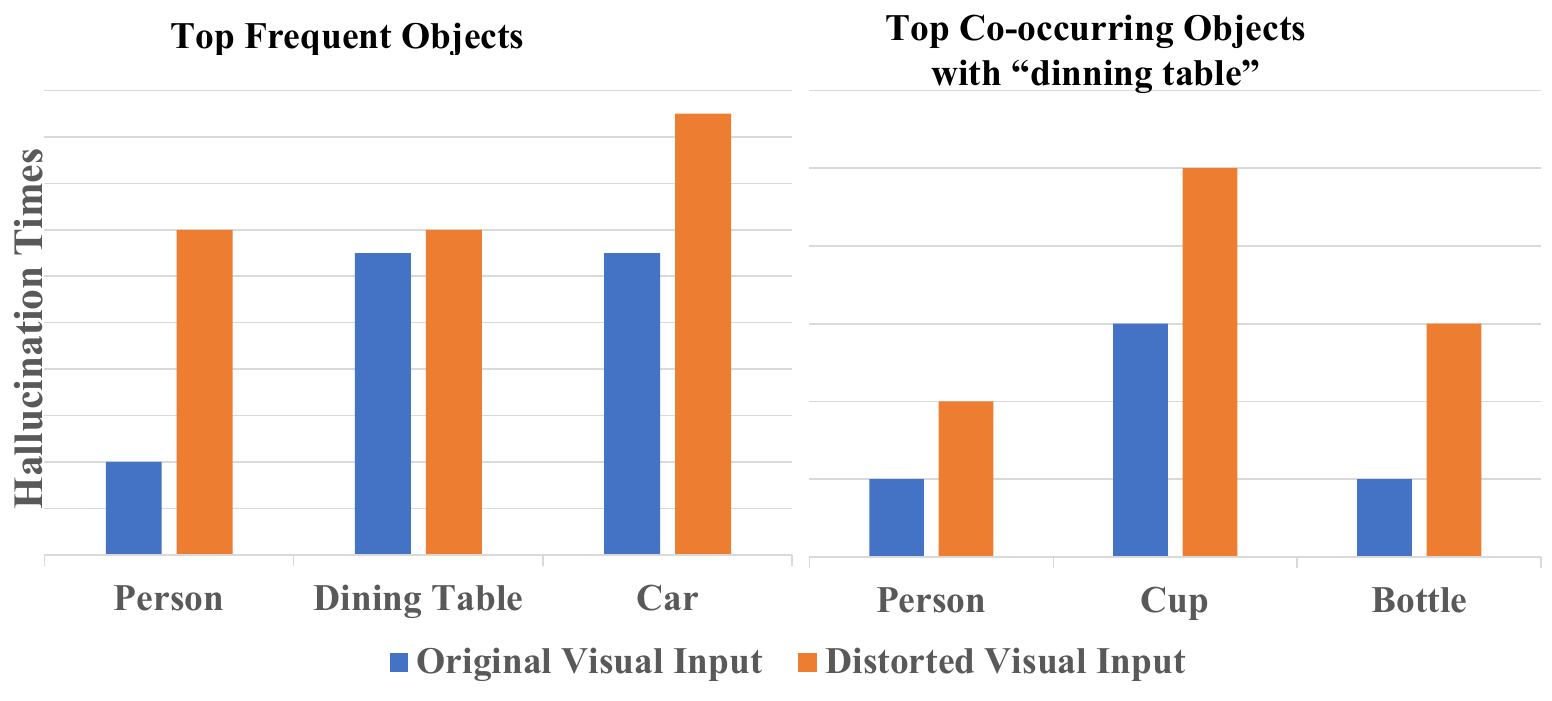}
    \caption{
    The left subfigure shows the correlation between frequent objects in MSCOCO and their propensity to be hallucinated in the validation set. Objects with a higher occurrence rate in the dataset are more likely to be hallucinated by LVLMs under distorted visual scenarios. 
    The right subfigure charts three objects that often appear alongside "\textit{dining table}", where they are also more frequently hallucinated when presented with distorted visual inputs.
    %Hallucination times of top frequent objects in MSCOCO and top co-occurring objects with \textit{"dining table"}.
    }
    \label{fig:bias}
    \vspace{-0.4cm}
\end{figure}

%% file: 04_experiments.tex
\section{Experiments}
\label{sec:experiments}
\input{tables/pope}

\input{tables/mme}
This section details our assessment of the proposed Visual Contrastive Decoding across various LVLMs.
\subsection{Experimental Settings}

\noindent\textbf{Datasets \& Evaluation Metrics}

\textbf{POPE}, the Polling-based Object Probing Evaluation \cite{li2023evaluating}, presents a streamlined approach to assess object hallucination.  
Within this benchmark, LVLMs are queried to answer if a specific object exists in the given image. The ratio between queries probing existent objects and non-existent objects is balanced (i.e.,$50$\% vs. $50$\%).
It encompasses three sampling settings: \textit{random, popular, and adversarial}, each distinct in constructing negative samples. In the \textit{random} setting, objects absent from the image are chosen randomly. The \textit{popular} setting selects missing objects from a high-frequency pool, while in the \textit{adversarial} setting, co-occurring objects not present in the image are prioritized.
The POPE benchmark aggregates data from three distinct sources: MSCOCO~\cite{lin2014microsoft}, A-OKVQA~\cite{schwenk2022okvqa}, and GQA~\cite{hudson2019gqa}.
It involves $500$ images from each dataset under each sampling setting and formulates $6$ questions per image, culminating in a total of $27,000$ query-answer pairs from the development sets of these datasets\footnote{Given the absence of ground-truth object annotations in A-OKVQA and GQA, SEEM~\cite{zou2023segment} is applied for image segmentation and object identification. }.
The evaluation pivots on four key metrics: Accuracy, Precision, Recall, and the F1 score.

%For each dataset, $3,000$ query-answer pairs are sampled across each sampling setting, totaling $9,000$ pairs from their dev sets.

%Here, Accuracy measures the fraction of queries addressed correctly. Precision and Recall respectively gauge the likelihood of accurate responses to ``\textit{Yes}" and ``\textit{No}" questions. The F1 score amalgamates the strengths of both Precision and Recall, emerging as the major metric for hallucination evaluation in this context.

\textbf{MME} \cite{fu2023mme} serves as an extensive benchmark tailored to assess LVLMs across multiple dimensions. It comprises ten perception-related subtasks and four cognition-focused ones. Following \citet{yin2023woodpecker}, except for adapting the whole dataset, we additionally leverage the existence and count subsets for object-level hallucination evaluation, and the position and color subsets for attribute-level hallucination assessment. 
%Mirroring POPE's configuration, each subset features “Yes-or-No” questions. 
Performance is quantified via the combined metric of accuracy and accuracy+ as the official implementation\footnote{\url{https://github.com/BradyFU/Awesome-Multimodal-Large-Language-Models/tree/Evaluation}}. 
%A higher score signifies enhanced performance and reduced hallucinations.

\textbf{LLaVA-Bench}\footnote{\url{https://huggingface.co/datasets/liuhaotian/llava-bench-in-the-wild}} features a collection of $24$ images, accompanying $60$ questions that span a range of contexts including indoor and outdoor scenes, memes, paintings, and sketches. This dataset is crafted to assess the capability of LVLMs in tackling more challenging tasks and their adaptability to new domains. We conduct case studies on this dataset to qualitatively demonstrate the effectiveness of our proposed VCD.

\vspace{0.2cm}
\noindent\textbf{LVLM Baselines}
We evaluate the effectiveness of our VCD on three state-of-the-art LVLMs. Concretely, we apply our VCD to LLaVA-1.5 and InstructBLIP, which employ Vicuna 7B as language decoder \cite{liu2023improved,dai2023instructblip}, 
and Qwen-VL, built on top of Qwen 7B backbone \cite{bai2023qwen}. 
For a more convincing comparison, we report the averaged results as well as the standard deviation over $5$ runs on POPE and MME benchmarks.
%Such a diverse selection of LVLMs provides a comprehensive platform to validate VCD's robustness across various architectures
%\lx{Note: the claim like this should be moved to results analysis, no need to say this here}.

\vspace{0.2cm}
\noindent\textbf{Implementation Details}
Throughout our experiments, we set $\alpha = 1$, $\beta = 0.1$, and $\gamma = 0.1$ unless explicitly stated otherwise. 
For a consistent comparative analysis, our baseline decoding strategy employs direct sampling (i.e., denoted as ``Regular" in all experimental tables), where the next token is directly sampled from the post-softmax distribution\footnote{Optimization of $\alpha$, $\beta$, $T$, and applying other sampling strategies as detailed in the ablation studies in Supplementary Materials may yield better results. The current settings serve as a constant baseline to demonstrate the efficacy of our approach.}. Conversely, instances labeled as``VCD" in the decoding column of all experimental tables refer to our proposed Visual Contrastive Decoding strategy, which also directly samples from the modified post-softmax distribution after applying VCD.
Comprehensive parameter configurations can be found in Supplementary Materials.
%For a more convincing comparison, we report the averaged results as well as the standard deviation over $5$ runs.
%Appendix~\ref{appendix:experimental settings}.

\subsection{Experimental Results}
\input{figs/chart1_1}

\noindent\textbf{Results on POPE}
Experimental results on POPE under the random, popular, and adversarial settings are summarized in Table~\ref{tab:pope}.
A notable observation is the robust effect of our proposed VCD. Specifically, under different sampling settings, the performances of our VCD consistently surpass the baseline results by large margins (up to +5.8 accuracy and +7.4 F1) on all of the LVLMs.
This suggests its pivotal role in counteracting statistical biases and language priors in LVLMs, thereby reducing instances of object hallucination.
In addition, all LVLMs display a clear performance degradation as we move from the \textit{random} setting to \textit{popular} and experience a further decline while moving to the \textit{adversarial} setting. This trend verifies our hypothesis that statistical biases inherent in LVLMs substantially contribute to the object hallucination problem. 
In a more detailed model-specific analysis, VCD demonstrates varied effects across different LVLMs. For LLaVA-1.5 and Qwen-VL, the F1 score elevation is predominantly driven by a recall boost (e.g., up to $10$ points), showcasing its enhanced ability to accurately detect object presences. Conversely, InstructBLIP's F1 score improvement is largely due to improved precision, signifying its enhanced capability to accurately filter out false positives. This highlights VCD's ability to accentuate distinct attributes of various model architectures in binary decision scenarios of POPE.

% Experimental results on POPE under the random, popular, and adversarial settings are summarized in Table~\ref{tab:pope}. A notable observation is the robust effect of our proposed VCD. Specifically, under different sampling settings, the performances of our VCD consistently surpass the baseline results by large margins (up to +5.8 accuracy and +7.4 F1) on all of the LVLMs. This suggests its pivotal role in counteracting statistical biases and language priors in LVLMs, thereby reducing instances of object hallucination. As can be seen, all LVLMs display a clear performance degradation as we move from the \textit{random} setting to \textit{popular} and experience a further decline while moving to the \textit{adversarial} setting. This trend verifies our hypothesis that statistical biases inherent in LVLMs substantially contribute to the object hallucination problem.

% dataset-wise interpretation
% Across MSCOCO, A-OKVQA, and GQA, we observe a consistent performance trend among models. Such consistency provides initial evidence for the reliability of the SEEM-based POPE. However, with all baseline models averaging F1 scores around $80$, there's a clear indication of the persistent hallucination challenges in LVLMs.
% Across three datasets, we observe consistent performances among models, with average F1 scores around $80$, indicating the persistent hallucination challenges in LVLMs.
% POPE-sampling-setting-wise interpretation

\input{figs/case}
\vspace{0.2cm}
\noindent\textbf{Results on MME Hallucination Subset}
The MME subset evaluations extend beyond POPE's scope, encompassing both object-level and attribute-level hallucinations. Results in Table~\ref{tab:mme} show that implementing VCD leads to a uniform enhancement in addressing object-level hallucinations for all models. 
% LLaVA-1.5 and Qwen-VL outperform InstructBLIP in object-level assessments—a trend consistent with POPE experiments. 
Additionally, VCD demonstrates an overall positive impact on attribute-level \textit{Color} scores, contributing to substantial overall performance gains. These improvements emphasize VCD's strength in addressing the embedded statistical bias and language priors of LVLMs, thus bringing a positive impact on a broader range of hallucination challenges. 
In contrast, the \textit{Position} score is relatively low across four metrics, with minimal uplift from VCD, suggesting the relatively weak ability of LVLMs in position reasoning.

\vspace{0.2cm}
\noindent\textbf{Results on MME Full Set}
As shown in Figure~\ref{chart:mme}, we also include the evaluation of VCD on MME Full Set to assess its impact on the general capability of LVLMs. With all models exhibiting comparable performance trajectories, we present the results of LLaVA-1.5 as a representative\footnote{Comprehensive results for all three LVLMs on the MME full set are provided in Supplementary Materials.}. 
%Appendix~\ref{appendix:mme}. 
The implementation of VCD leads to a consistent enhancement in perception-based tasks, while the original recognition competencies of the LVLMs are preserved. 
This may be attributed to VCD’s reduction of statistical bias and language priors, which improves LVLMs' general perception capacities by ensuring a visually grounded analysis.
%This may be attributed to VCD's effective mitigation of the embedded statistical bias and language priors within LVLMs. 
%In perception-related tasks such as identifying landmarks in images, VCD’s reduction of statistical bias and language priors encourages LVLMs to prioritize visual data over textual expectations, improving general perception capacities by ensuring a visually grounded analysis.
%For example, when discerning the ``\textit{Eiffel Tower}" in a cluttered scene, LVLMs without VCD may default to frequent co-occurrences like ``\textit{Paris}" or ``\textit{cityscape}". VCD, by diminishing these priors, compels the model to focus on the distinctive structural features of the landmark, improving task performance by ensuring a visually grounded analysis.

%Due to the page limit, please refer to Supplementary Materials for additional case studies and ablation studies.

\subsection{Further Discussions}
\input{figs/noisy_prior}
\noindent\textbf{Effect of Visual Uncertainty on Hallucinations}
%Figure~\ref{fig: noisy_prior} presents an ablation study examining the impact of varying noise steps, denoted as $T$, using the LLaVA-1.5 model with regular decoding on the POPE benchmark. 
We further study how the object hallucination of LLaVA-1.5 changes along with visual uncertainty. Figure~\ref{fig: noisy_prior} depicts a clear performance drop on the POPE benchmark with the increase of noise steps, suggesting that the object hallucination will become more severe as visual uncertainty goes larger. This observation aligns with our previous findings in Section~\ref{subsec: observation} that visual uncertainty will exacerbate object hallucination issues in LVLMs' generative process. Our proposed VCD emerges as a correction mechanism by contrasting model outputs with original and distorted visual inputs.

\vspace{0.1cm}
\noindent\textbf{GPT-4V Aided Evaluation of Open-Ended Generation}
% enhance perception because of mitigation of language priors and statistical biases.
Beyond the ``Yes-or-No" question format employed in our POPE and MME evaluations, we extend our analysis to open-ended captioning tasks in the LLaVA-Bench using the recently released LVLM, GPT-4V\footnote{\url{https://openai.com/research/gpt-4v-system-card}}, following \citet{yin2023woodpecker}\footnote{The prompt used for evaluation and an evaluation case is provided in Supplementary Materials.}. 
%Following \citet{yin2023woodpecker}, we utilize two key metrics for evaluation: (1) accuracy, measuring the response's alignment with the image content, and (2) detailedness, gauging the richness of details in the response. 
Results in Table~\ref{tab:gpt4v} show consistent improvements in VCD over regular decoding. 
The observed enhancement in accuracy points to VCD's ability to mitigate hallucinations effectively. Simultaneously, VCD's counteraction of statistical biases and language priors enhances the perceptual capabilities of LVLMs, as evidenced by the marked improvement in the detailedness of the responses.

\vspace{0.2cm}
\noindent\textbf{Case Study on LLaVA-Bench}
Figure~\ref{fig:case study} demonstrates two case studies on how, given identical prompts and images, regular decoding can yield object hallucinations influenced by the statistical bias and language priors inherent during pretraining. For instance, in the displayed examples, objects such as ``\textit{dining table}" and ``\textit{fork}", which often co-occur with the likely ground-truth object ``\textit{chair}", are hallucinated.
In contrast, the implementation of VCD notably mitigates these hallucination issues and simultaneously preserves the coherence and informativeness of the output text. Due to the page limit, please refer to Supplementary Materials for more cases and ablation studies\footnote{Ablation studies in Supplementary Materials include effects of total noise steps $T$, hyper-parameters $\alpha$, $\beta$, and effect of VCD on larger LVLM variants and with other sampling strategies.}.

\input{tables/gpt4}

%% file: tables/pope.tex
\begin{table*}[tp]
\centering
\resizebox{0.9\linewidth}{!}{%
\begin{tabular}{cllllll|l}
\hline
\textbf{Dataset}          & \textbf{Setting}                         & \textbf{Model}                & \textbf{Decoding} & Accuracy$\uparrow$ & Precision & Recall & F1 Score$\uparrow$  \\ \hline
\multirow{18}{*}{MSCOCO}  & \multirow{6}{*}{\textit{Random}}      & \multirow{2}{*}{LLaVA1.5}     & Regular           &$83.29_{(\pm0.35)}$ &$92.13_{(\pm0.54)}$ &$72.80_{(\pm0.57)}$ &$81.33_{(\pm0.41)}$                                         \\
                          &                                       &                               & VCD               &$\textbf{87.73}_{(\pm0.40)}$ &$91.42_{(\pm0.55)}$ &$83.28_{(\pm0.42)}$ &$\textbf{87.16}_{(\pm0.41)}$  \\
                          &                                       & \multirow{2}{*}{Qwen-VL}     & Regular           &$84.73_{(\pm0.36)}$ &$95.61_{(\pm0.45)}$ &$72.81_{(\pm0.38)}$ &$82.67_{(\pm0.41)}$  \\
                          &                                       &                               & VCD               &$\textbf{88.63}_{(\pm0.10)}$ &$94.64_{(\pm0.25)}$ &$81.91_{(\pm0.19)}$ &$\textbf{87.81}_{(\pm0.11)}$  \\
                          &                                       & \multirow{2}{*}{InstructBLIP} & Regular           &$80.71_{(\pm0.73)}$ &$81.67_{(\pm0.67)}$ &$79.19_{(\pm1.14)}$ &$80.41_{(\pm0.80)}$  \\
                          &                                       &                               & VCD               &$\textbf{84.53}_{(\pm0.38)}$ &$88.55_{(\pm0.54)}$ &$79.32_{(\pm0.44)}$ &$\textbf{83.68}_{(\pm0.40)}$  \\ \cline{2-8} 
                          & \multirow{6}{*}{\textit{Popular}}     & \multirow{2}{*}{LLaVA1.5}     & Regular           &$81.88_{(\pm0.48)}$ &$88.93_{(\pm0.60)}$ &$72.80_{(\pm0.57)}$ &$80.06_{(\pm0.05)}$  \\
                          &                                       &                               & VCD               &$\textbf{85.38}_{(\pm0.38)}$ &$86.92_{(\pm0.53)}$ &$83.28_{(\pm0.42)}$ &$\textbf{85.06}_{(\pm0.37)}$   \\
                          &                                       & \multirow{2}{*}{Qwen-VL}     & Regular           &$84.13_{(\pm0.18)}$ &$94.31_{(\pm0.43)}$ &$72.64_{(\pm0.45)}$ &$82.06_{(\pm0.23)}$  \\
                          &                                       &                               & VCD               &$\textbf{87.12}_{(\pm0.07)}$ &$91.49_{(\pm0.10)}$ &$81.85_{(\pm0.19)}$ &$\textbf{86.40}_{(\pm0.09)}$  \\
                          &                                       & \multirow{2}{*}{InstructBLIP} & Regular           &$78.22_{(\pm0.84)}$ &$77.87_{(\pm1.03)}$ &$78.85_{(\pm0.52)}$ &$78.36_{(\pm0.76)}$  \\
                          &                                       &                               & VCD               &$\textbf{81.47}_{(\pm0.42)}$ &$82.89_{(\pm0.64)}$ &$79.32_{(\pm0.44)}$ &$\textbf{81.07}_{(\pm0.39)}$  \\ \cline{2-8} 
                          & \multirow{6}{*}{\textit{Adversarial}} & \multirow{2}{*}{LLaVA1.5}     & Regular           &$78.96_{(\pm0.52)}$ &$83.06_{(\pm0.58)}$ &$72.75_{(\pm0.59)}$ &$77.57_{(\pm0.57)}$  \\
                          &                                       &                               & VCD               &$\textbf{80.88}_{(\pm0.33)}$ &$79.45_{(\pm0.29)}$ &$83.29_{(\pm0.43)}$ &$\textbf{81.33}_{(\pm0.34)}$  \\
                          &                                       & \multirow{2}{*}{Qwen-VL}     & Regular           &$82.26_{(\pm0.30)}$ &$89.97_{(\pm0.33)}$ &$72.61_{(\pm0.50)}$ &$80.37_{(\pm0.37)}$  \\
                          &                                       &                               & VCD               &$\textbf{84.26}_{(\pm0.39)}$ &$85.84_{(\pm0.45)}$ &$82.05_{(\pm0.39)}$ &$\textbf{83.90}_{(\pm0.39)}$  \\
                          &                                       & \multirow{2}{*}{InstructBLIP} & Regular           &$75.84_{(\pm0.45)}$ &$74.30_{(\pm0.63)}$ &$79.03_{(\pm0.68)}$ &$76.59_{(\pm0.40)}$  \\
                          &                                       &                               & VCD               &$\textbf{79.56}_{(\pm0.41)}$ &$79.67_{(\pm0.59)}$ &$79.39_{(\pm0.50)}$ &$\textbf{79.52}_{(\pm0.38)}$  \\ \hline
\multirow{18}{*}{A-OKVQA} & \multirow{6}{*}{\textit{Random}}      & \multirow{2}{*}{LLaVA1.5}     & Regular           &$83.45_{(\pm0.48)}$ &$87.24_{(\pm0.68)}$ &$78.36_{(\pm0.54)}$ &$82.56_{(\pm0.50)}$  \\
                          &                                       &                               & VCD               &$\textbf{86.15}_{(\pm0.23)}$ &$85.18_{(\pm0.34)}$ &$87.53_{(\pm0.14)}$ &$\textbf{86.34}_{(\pm0.21)}$  \\
                          &                                       & \multirow{2}{*}{Qwen-VL}     & Regular           &$86.67_{(\pm0.48)}$ &$93.16_{(\pm0.55)}$ &$79.16_{(\pm0.59)}$ &$85.59_{(\pm0.53)}$  \\
                          &                                       &                               & VCD               &$\textbf{89.22}_{(\pm0.14)}$ &$90.77_{(\pm0.04)}$ &$87.32_{(\pm0.34)}$ &$\textbf{89.01}_{(\pm0.16)}$  \\
                          &                                       & \multirow{2}{*}{InstructBLIP} & Regular           &$80.91_{(\pm0.34)}$ &$77.97_{(\pm0.59)}$ &$86.16_{(\pm0.88)}$ &$81.86_{(\pm0.32)}$  \\
                          &                                       &                               & VCD               &$\textbf{84.11}_{(\pm0.27)}$ &$82.21_{(\pm0.35)}$ &$87.05_{(\pm0.53)}$ &$\textbf{84.56}_{(\pm0.28)}$  \\ \cline{2-8} 
                          & \multirow{6}{*}{\textit{Popular}}     & \multirow{2}{*}{LLaVA1.5}     & Regular           &$79.90_{(\pm0.33)}$ &$80.85_{(\pm0.31)}$ &$78.36_{(\pm0.54)}$ &$79.59_{(\pm0.37)}$  \\
                          &                                       &                               & VCD               &$\textbf{81.85}_{(\pm0.44)}$ &$78.60_{(\pm0.58)}$ &$87.53_{(\pm0.14)}$ &$\textbf{82.82}_{(\pm0.36)}$  \\
                          &                                       & \multirow{2}{*}{Qwen-VL}     & Regular           &$85.56_{(\pm0.35)}$ &$90.44_{(\pm0.56)}$ &$79.53_{(\pm0.84)}$ &$84.63_{(\pm0.42)}$  \\
                          &                                       &                               & VCD               &$\textbf{87.85}_{(\pm0.30)}$ &$88.10_{(\pm0.36)}$ &$87.53_{(\pm0.47)}$ &$\textbf{87.81}_{(\pm0.31)}$  \\
                          &                                       & \multirow{2}{*}{InstructBLIP} & Regular           &$76.19_{(\pm0.80)}$ &$72.16_{(\pm0.69)}$ &$85.28_{(\pm0.79)}$ &$78.17_{(\pm0.73)}$  \\
                          &                                       &                               & VCD               &$\textbf{79.78}_{(\pm0.47)}$ &$76.00_{(\pm0.52)}$ &$87.05_{(\pm0.53)}$ &$\textbf{81.15}_{(\pm0.42)}$  \\ \cline{2-8} 
                          & \multirow{6}{*}{\textit{Adversarial}} & \multirow{2}{*}{LLaVA1.5}     & Regular           &$74.04_{(\pm0.34)}$ &$72.08_{(\pm0.53)}$ &$78.49_{(\pm0.38)}$ &$75.15_{(\pm0.23)}$  \\
                          &                                       &                               & VCD               &$\textbf{74.97}_{(\pm0.39)}$ &$70.01_{(\pm0.40)}$ &$87.36_{(\pm0.15)}$ &$\textbf{77.73}_{(\pm0.29)}$  \\
                          &                                       & \multirow{2}{*}{Qwen-VL}     & Regular           &$79.57_{(\pm0.31)}$ &$79.77_{(\pm0.34)}$ &$79.23_{(\pm0.73)}$ &$79.50_{(\pm0.38)}$  \\
                          &                                       &                               & VCD               &$\textbf{81.27}_{(\pm0.09)}$ &$77.79_{(\pm0.20)}$ &$87.53_{(\pm0.34)}$ &$\textbf{82.38}_{(\pm0.10)}$  \\
                          &                                       & \multirow{2}{*}{InstructBLIP} & Regular           &$70.71_{(\pm0.76)}$ &$65.91_{(\pm0.74)}$ &$85.83_{(\pm0.80)}$ &$75.56_{(\pm0.57)}$  \\
                          &                                       &                               & VCD               &$\textbf{74.33}_{(\pm0.67)}$ &$69.46_{(\pm0.73)}$ &$86.87_{(\pm0.27)}$ &$\textbf{77.19}_{(\pm0.47)}$  \\ \hline
\multirow{18}{*}{GQA}     & \multirow{6}{*}{\textit{Random}}      & \multirow{2}{*}{LLaVA1.5}     & Regular           &$83.73_{(\pm0.27)}$ &$87.16_{(\pm0.39)}$ &$79.12_{(\pm0.35)}$ &$82.95_{(\pm0.28)}$  \\
                          &                                       &                               & VCD               &$\textbf{86.65}_{(\pm0.45)}$ &$84.85_{(\pm0.59)}$ &$89.24_{(\pm0.34)}$ &$\textbf{86.99}_{(\pm0.41)}$  \\
                          &                                       & \multirow{2}{*}{Qwen-VL}     & Regular           &$80.97_{(\pm0.32)}$ &$88.07_{(\pm0.34)}$ &$71.64_{(\pm0.57)}$ &$79.01_{(\pm0.40)}$  \\
                          &                                       &                               & VCD               &$\textbf{85.59}_{(\pm0.38)}$ &$86.88_{(\pm0.44)}$ &$83.84_{(\pm0.36)}$ &$\textbf{85.33}_{(\pm0.38)}$  \\
                          &                                       & \multirow{2}{*}{InstructBLIP} & Regular           &$79.65_{(\pm0.24)}$ &$77.14_{(\pm0.43)}$ &$84.29_{(\pm0.36)}$ &$80.56_{(\pm0.18)}$  \\
                          &                                       &                               & VCD               &$\textbf{83.69}_{(\pm0.11)}$ &$81.84_{(\pm0.42)}$ &$86.61_{(\pm0.48)}$ &$\textbf{84.16}_{(\pm0.01)}$  \\ \cline{2-8} 
                          & \multirow{6}{*}{\textit{Popular}}     & \multirow{2}{*}{LLaVA1.5}     & Regular           &$78.17_{(\pm0.17)}$ &$77.64_{(\pm0.26)}$ &$79.12_{(\pm0.35)}$ &$78.37_{(\pm0.18)}$  \\
                          &                                       &                               & VCD               &$\textbf{80.73}_{(\pm0.47)}$ &$76.26_{(\pm0.68)}$ &$89.24_{(\pm0.34)}$ &$\textbf{82.24}_{(\pm0.35)}$  \\
                          &                                       & \multirow{2}{*}{Qwen-VL}     & Regular           &$75.99_{(\pm0.33)}$ &$78.62_{(\pm0.41)}$ &$71.40_{(\pm0.38)}$ &$74.84_{(\pm0.34)}$  \\
                          &                                       &                               & VCD               &$\textbf{81.83}_{(\pm0.27)}$ &$80.45_{(\pm0.47)}$ &$84.09_{(\pm0.32)}$ &$\textbf{82.23}_{(\pm0.22)}$  \\
                          &                                       & \multirow{2}{*}{InstructBLIP} & Regular           &$73.87_{(\pm0.58)}$ &$69.63_{(\pm0.54)}$ &$84.69_{(\pm0.68)}$ &$76.42_{(\pm0.52)}$  \\
                          &                                       &                               & VCD               &$\textbf{78.57}_{(\pm0.14)}$ &$74.62_{(\pm0.22)}$ &$86.61_{(\pm0.48)}$ &$\textbf{80.17}_{(\pm0.16)}$  \\ \cline{2-8} 
                          & \multirow{6}{*}{\textit{Adversarial}} & \multirow{2}{*}{LLaVA1.5}     & Regular           &$75.08_{(\pm0.33)}$ &$73.19_{(\pm0.49)}$ &$79.16_{(\pm0.35)}$ &$76.06_{(\pm0.24)}$  \\
                          &                                       &                               & VCD               &$\textbf{76.09}_{(\pm0.43)}$ &$70.83_{(\pm0.45)}$ &$88.75_{(\pm0.56)}$ &$\textbf{78.78}_{(\pm0.36)}$  \\
                          &                                       & \multirow{2}{*}{Qwen-VL}     & Regular           &$75.46_{(\pm0.63)}$ &$77.92_{(\pm0.73)}$ &$71.07_{(\pm0.97)}$ &$74.33_{(\pm0.71)}$  \\
                          &                                       &                               & VCD               &$\textbf{80.01}_{(\pm0.27)}$ &$77.86_{(\pm0.24)}$ &$83.85_{(\pm0.35)}$ &$\textbf{80.75}_{(\pm0.27)}$  \\
                          &                                       & \multirow{2}{*}{InstructBLIP} & Regular           &$70.56_{(\pm0.53)}$ &$66.12_{(\pm0.32)}$ &$84.33_{(\pm1.05)}$ &$74.12_{(\pm0.58)}$  \\
                          &                                       &                               & VCD               &$\textbf{75.08}_{(\pm0.13)}$ &$70.59_{(\pm0.16)}$ &$85.99_{(\pm0.10)}$ &$\textbf{77.53}_{(\pm0.08)}$  \\ \hline
\end{tabular}
}
\caption{Results on POPE. \textit{Regular} decoding denotes direct sampling, whereas \textit{VCD} refers to sampling from our proposed contrastive distribution $p_{vcd}$. 
%Higher accuracy and F1 score indicate better performance and fewer hallucinations. 
The best performances within each setting are \textbf{bolded}.}
\label{tab:pope}
\end{table*}

%% file: tables/mme.tex
\begin{table*}[h]
\centering
\resizebox{0.825\linewidth}{!}{%
\begin{tabular}{@{}lllllll@{}}
\toprule
\multirow{2}{*}{Model}        & \multirow{2}{*}{Decoding} & \multicolumn{2}{c}{\textbf{Object-level}}                                   & \multicolumn{2}{c}{\textbf{Attribute-level}}                               & \multicolumn{1}{c}{\multirow{2}{*}{Total Scores$\uparrow$}} \\
                              &                           & \multicolumn{1}{c}{\textit{Existence}$\uparrow$} & \multicolumn{1}{c}{\textit{Count}$\uparrow$} & \multicolumn{1}{c}{\textit{Position}$\uparrow$} & \multicolumn{1}{c}{\textit{Color}$\uparrow$} & \multicolumn{1}{c}{}                       \\ \midrule
\multirow{2}{*}{LLaVA1.5}     & Regular                   &$175.67_{(\pm7.51)}$ &$124.67_{(\pm19.59)}$ &$114.00_{(\pm9.32)}$ &$151.00_{(\pm10.45)}$ &$565.33_{(\pm32.92)}$ \\
                              & VCD                       &$\textbf{184.66}_{(\pm6.81)}$ &$\textbf{138.33}_{(\pm15.68)}$ &$\textbf{128.67}_{(\pm7.21)}$ &$\textbf{153.00}_{(\pm7.58)}$ &$\textbf{604.66}_{(\pm18.76)}$ \\ \midrule
\multirow{2}{*}{Qwen-VL}   & Regular                   &$155.00_{(\pm3.54)}$ &$127.67_{(\pm13.36)}$ &$\textbf{131.67}_{(\pm7.73)}$ &$173.00_{(\pm9.75)}$ &$587.33_{(\pm31.06)}$ \\
                              & VCD                       &$\textbf{156.00}_{(\pm6.52)}$ &$\textbf{131.00}_{(\pm6.19)}$ &$128.00_{(\pm3.61)}$ &$\textbf{181.67}_{(\pm5.14)}$ &$\textbf{596.67}_{(\pm11.61)}$ \\ \midrule
\multirow{2}{*}{InstructBLIP} & Regular                   &$141.00_{(\pm13.97)}$ &$75.33_{(\pm14.16)}$ &$\textbf{66.67}_{(\pm3.91)}$ &$97.33_{(\pm16.94)}$ &$380.33_{(\pm40.20)}$ \\
                              & VCD                       &$\textbf{168.33}_{(\pm11.55)}$ &$\textbf{92.33}_{(\pm8.47)}$ &$64.00_{(\pm6.73)}$ &$\textbf{123.00}_{(\pm11.27)}$ &$\textbf{447.67}_{(\pm13.36)}$ \\ \bottomrule
\end{tabular}
}
\caption{Results on the hallucination subset of MME. Regular decoding denotes direct sampling, whereas VCD refers to sampling from our proposed contrastive distribution $p_{vcd}$. 
%Higher scores indicate better performance and fewer hallucinations. 
The best performances within each setting are \textbf{bolded}.}
\label{tab:mme}
%\vspace{-0.2cm}
\end{table*}

%% file: figs/chart1_1.tex
\begin{figure*}[tp]
    \centering
    \includegraphics[width=0.95\linewidth]{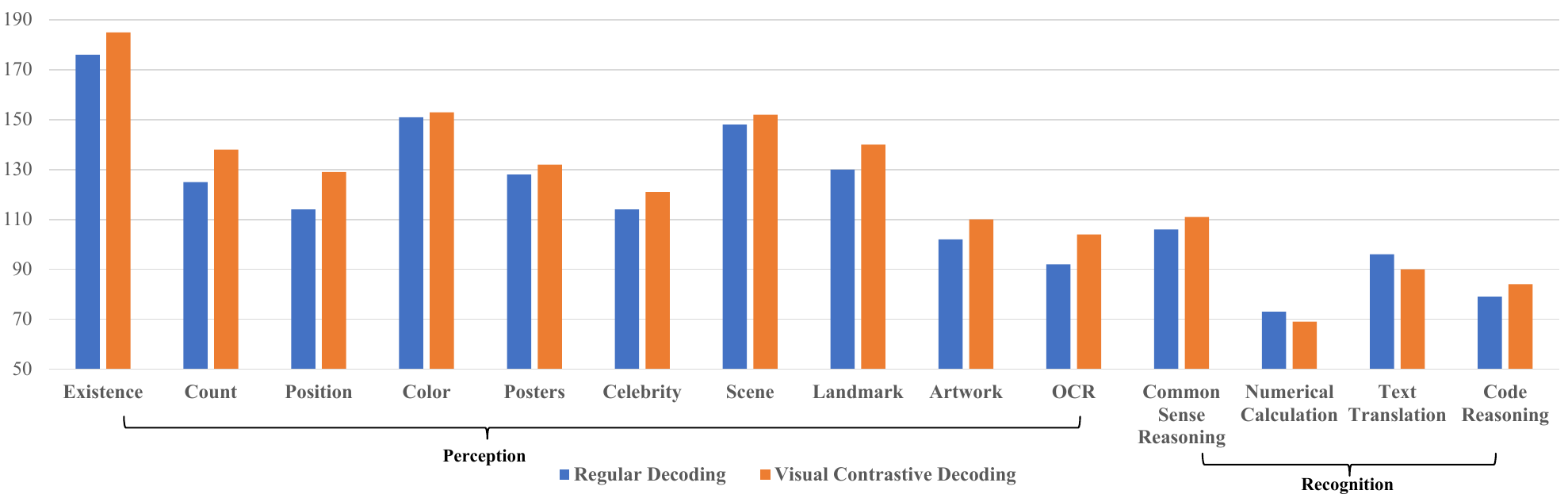}
    \caption{MME full set results on LLaVA-1.5. VCD leads to consistent enhancement in LVLMs' perception capacities while preserving their recognition competencies.}
    \label{chart:mme}
    % \vspace{-0.2cm}
\end{figure*}

%% file: figs/case.tex
\begin{figure*}[tp]
    \centering
    \includegraphics[width=1\linewidth]{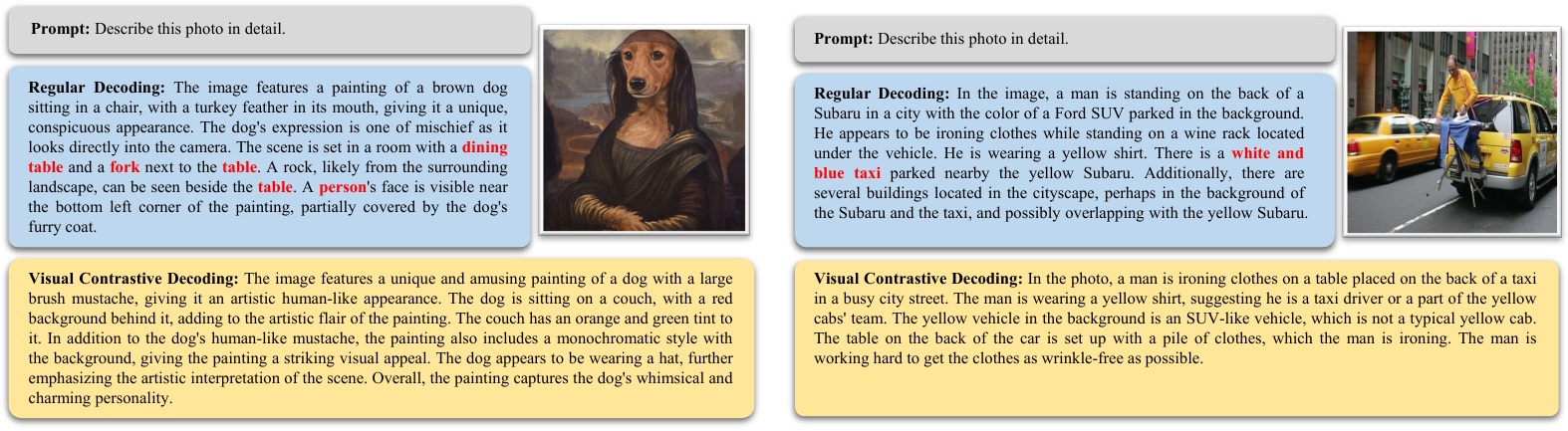}
    \caption{Illustration of hallucination correction by our proposed VCD with two samples from LLaVA-Bench. Hallucinated objects from LVLM's regular decoding are highlighted in \color{red}{red}.}
    \label{fig:case study}
    \vspace{-0.1cm}
\end{figure*}

%% file: figs/noisy_prior.tex
\begin{figure}[tp]
    \centering
    \includegraphics[width=0.85\linewidth]{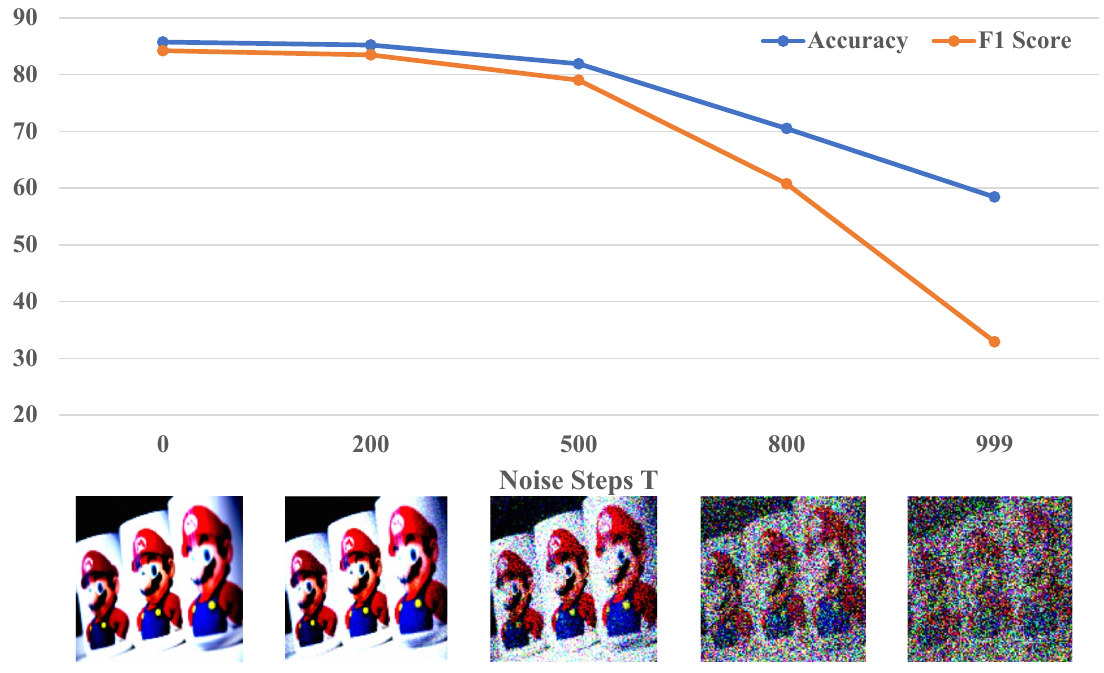}
    \caption{Performance of LLaVA-1.5 on the POPE benchmark across varying noise levels with regular decoding. We visualize the distorted visual inputs subjected to different levels of Gaussian noise at the bottom.}
    \label{fig: noisy_prior}
    %\vspace{-0.2cm}
\end{figure}

%% file: tables/gpt4.tex
\begin{table}[tp]
\centering
\resizebox{0.8\linewidth}{!}{%
\begin{tabular}{@{}llcc@{}}
\toprule
\textbf{Model}                & \textbf{Decoding} & Accuracy$\uparrow$ & Detailedness$\uparrow$ \\ \midrule
\multirow{2}{*}{LLaVA-1.5}     & Regular           & $3.23$         & $3.54$            \\
                              & VCD               & $\textbf{4.15}$         & $\textbf{3.85}$            \\ \midrule
\multirow{2}{*}{InstructBLIP}      & Regular           &   $3.84$       &       $4.07$      \\
                              & VCD               &   $\textbf{4.23}$       &     $\textbf{4.69}$        \\ \midrule
\multirow{2}{*}{Qwen-VL} & Regular           &      $4.76$    &    $3.46$         \\
                              & VCD               &       $\textbf{6.69}$   &    $\textbf{4.46}$       \\ \bottomrule
\end{tabular}
}
\caption{Results of GPT-4V-aided evaluation on open-ended generation.
Accuracy measures the response's alignment with the image content, and Detailedness gauges the richness of details in the response.
Both metrics are on a scale of 10.}
\label{tab:gpt4v}
%\vspace{-0.2cm}
\end{table}

%% file: 10_conclusion.tex
\section{Conclusion and Limitation}
\label{sec:conclusion}
In this paper, we tackle the object hallucination issue in LVLMs. 
We conducted an in-depth analysis of how visual uncertainty influences hallucinations, particularly from the aspect of statistical biases and language priors. Our findings indicate that visual uncertainty amplifies these factors, contributing to more hallucinations.
In light of this, we introduced Visual Contrastive Decoding (VCD), a novel, training-free method that employs contrastive distributions to calibrate the model's output without the usage of external tools. Our extensive experiments across multiple benchmarks and LVLM families confirm VCD's efficacy in reducing hallucinations and also demonstrate its potential to enhance the overall perception capabilities of LVLMs. 

\vspace{0.2cm}
\noindent\textbf{Limitation}
% While this study employs a basic Gaussian noise approach, more fine-grained techniques, like object-level blurring, hold the potential for improved outcomes. Exploring such distortions is considered a direction for future research.
While this study employs a basic Gaussian noise approach to introduce visual uncertainty, more fine-grained techniques, like object-level blurring, hold the potential for improved outcomes. In addition, our focus was limited to LVLMs processing images and text, not encompassing their emerging applications in video understanding. Future research directions include exploring diverse image distortion methods and extending the Visual Contrastive Decoding (VCD) framework to a broader range of LVLMs.

%% file: 12_appendix.tex
\section{Detailed Experimental Settings}
\label{appendix:experimental settings}
In all experimental setups, the hyper-parameters $\gamma$, $\alpha$ and $\beta$, as specified in Equations~\ref{eq:1}, \ref{eq:3} and \ref{eq:5}, are fixed at values of $0.1$, $1$ and $0.1$, respectively. For the total number of noise steps $T$ delineated in Equation~\ref{eq:1}, we set a value of $500$ for experiments involving the MME and LLaVA-Bench, while for those evaluating on POPE, the $T$ value is set at $999$.

\section{Ablation Studies}
\label{appendix:ablation}
For the Ablation Studies section, the default configuration for hyper-parameters $\alpha$, $\beta$, and $\delta$ is set to $1$, $0.1$, and $500$, respectively. These values are retained across all experiments unless an individual study specifies an alternative parameter adjustment for investigation. Across all the experiments, we use LLaVA-1.5 as the representative LVLM baseline to demonstrate the effect of tuning different hyper-parameters.
\subsection{Effect of Total Noise Steps $T$}
\input{tables/ablation1}
Figure~\ref{tab:ablation1} presents an ablation study examining the impact of varying noise levels, denoted as $\delta$, using the LLaVA-1.5 model on the MME benchmark. In alignment with the experimental configuration, MME is subdivided into three subsets: hallucination, perception, and recognition. The hallucination subset includes tasks related to \textit{Existence}, \textit{Count}, \textit{Position}, and \textit{Color}, while the perception subset encompasses these and additional perception-focused tasks. The recognition subset, conversely, involves tasks that challenge LVLMs' cognitive reasoning abilities.

The study reveals a pronounced sensitivity to different $\delta$ values within the hallucination subset, where optimal noise levels correlate with substantially enhanced overall scores. In the realm of perception tasks, a surpassing of a specific noise threshold ($\delta > 500$) showcases VCD's capability to consistently yield improvements. For recognition tasks, VCD maintains steady performance across the spectrum of tested noise values.

\subsection{Effect of $\alpha$ in Visual Contrastive Decoding }
\input{tables/ablation2}
Table~\ref{tab:ablation2} demonstrates the outcomes of an ablation study focusing on $\alpha$, which modulates the level of amplification between output distributions from original and distorted visual inputs, as formulated in Equation~\ref{eq:3}. The study observes minimal variance in the aggregate scores across the three MME subsets as $\alpha$ ranges from $0.25$ to $1.0$, showcasing a uniform improvement over regular decoding. This consistency evidences the efficacy and stability of the contrastive decoding strategy across a spectrum of $\alpha$ settings.

\subsection{Effect of $\beta$ in Adaptive Plausible Constraint }
\input{tables/ablation3}
Table~\ref{tab:ablation3} presents the results of an ablation study on $\beta$, which controls the adaptive plausible constraint in Equation~\ref{eq:5}, where larger $\beta$ indicates more aggressive truncation, keeping only high-probability tokens. The table illustrates that a $\beta$ value of $0$, implying no constraint, results in suboptimal performance, which validates our rationale for implementing this constraint: the output distribution with distorted visual inputs can still uphold fundamental linguistic standards and common sense reasoning. Indiscriminate penalization could inadvertently sanction these valid outputs and promote the generation of implausible tokens. 
As $\beta$ increases, improvements in total scores across the hallucination and perception subsets are observed, highlighting the constraint's critical role in reducing hallucinations and improving LVLMs' perception capacities. 
%Notably, a $\beta$ value of $1$ yields the highest performance, suggesting that limiting the selection to one or two high-probability tokens, approaching to the greedy decoding, leads to the best results. This outcome underscores the LVLMs' heightened confidence after applying VCD and affirms VCD's capacity to robustly address hallucinations and augment broader LVLM capabilities, even within the most assertive decoding strategy, like greedy decoding.

\subsection{Effect of Different Sampling Strategies}
\input{tables/ablation4}
Table~\ref{tab:ablation4} presents an ablation study on various sampling strategies conducted on the POPE-\textit{Random} dataset using LLaVA-1.5.  In addition to the direct sampling approach discussed in the main paper, this experiment includes four additional sampling strategies: Top P sampling (specifically, $p=0.9$), Top K sampling (specifically, $k=50$), Greedy decoding, and Top K sampling with temperature normalization ($k=50, temp=1.5/0.7$). The results indicate that applying VCD, irrespective of the sampling strategy employed, consistently contributes to hallucination mitigation and an enhancement of the general performance capabilities of LVLMs. This consistency underscores the versatility and effectiveness of VCD across different sampling strategies in the context of LVLMs.

\subsection{Effect of VCD when LVLMs Scale Up}
\input{tables/pope_upscale}
Our evaluation extends to larger 13B variants of the LLaVA-1.5 and InstructBLIP models\footnote{Qwen-VL lacks larger variants.}, assessing the scalability of our proposed VCD across different LVLM magnitudes. Table~\ref{tab: pope_upscale} reveals that the 7B and 13B variants of LLaVA-1.5 and InstructBLIP exhibit comparable performances across POPE settings (e.g., $81.33$ and $81.49$ F1 scores for LLaVA-1.5 7B and 13B in \textit{Random} setting), suggesting that increasing the model parameters does not inherently resolve hallucination issues, thereby underscoring the pertinence of addressing this challenge. Crucially, VCD consistently boosts performance in all POPE configurations, reaffirming its robustness independent of model scale.

\section{Detailed Experimental Results on MME}
\label{appendix:mme}
\input{tables/mme_percep}
In Table~\ref{tab:mme_perception}, we present the performance of three LVLM baselines on the perception-related tasks of the MME benchmark. The baselines exhibit consistent performance patterns, and the deployment of VCD uniformly improves their perceptual competencies. This improvement is likely a consequence of VCD's capability to diminish statistical biases and language priors, thus recalibrating the LVLMs to favor visual information over pre-existing biases and priors.

\input{tables/mme_recog}
Furthermore, Table~\ref{tab:mme_recognition} showcases the LVLMs' performances on recognition-related tasks within the MME benchmark. The results indicate that the application of VCD, while alleviating hallucination issues and augmenting perceptual capabilities, does not compromise the inherent reasoning abilities of LVLMs, as evidenced by the stable overall recognition scores.

\section{More Case Studies}
\label{appendix:more case studies}
\input{figs/case_hallu}
\input{figs/case_general}
Additional case studies on the LLaVA-bench are presented to illustrate the effectiveness of our approach across different LVLMs. Figure~\ref{fig:case_hallu} provides further instances of hallucination corrections by our method. Meanwhile, Figure~\ref{fig:case_general} offers supplemental examples of the enhancements brought by our proposed VCD in bolstering the general perception and recognition abilities of LVLMs.

\input{tables/GPT4_prompt}
\input{figs/gpt4_evaluator_case}
\section{Prompt and Case for GPT-4V Aided Evaluation}
To evaluate open-ended generation, we utilize GPT-4V to assess the accuracy and detailedness of LVLMs' responses.
The specific configurations are detailed in Table~\ref{tab:prompt_evaluation}. Additionally, an illustrative evaluation case is presented in Figure~\ref{fig: gpt4_evaluator_case}.

%% file: tables/ablation1.tex
\begin{table}[t]
\resizebox{1\linewidth}{!}{%
\begin{tabular}{@{}c|ccc@{}}
\toprule
\textbf{$T$} & \multicolumn{1}{c}{\textit{\begin{tabular}[c]{@{}c@{}}Hallucination Subset\\ Total Scores\end{tabular}}} & \multicolumn{1}{c}{\textit{\begin{tabular}[c]{@{}c@{}}Perception Subset\\ Total Scores\end{tabular}}} & \multicolumn{1}{c}{\textit{\begin{tabular}[c]{@{}c@{}}Recognition Subset\\ Total Scores\end{tabular}}} \\ \midrule
200                  & $586.67_{\pm 11.67}$                                                     & $1311.47_{\pm 4.33}$                                                     & $338.69_{\pm 19.87}$                                                                             \\
500                  & $591.67_{\pm 36.06}$                                                     & $1340.89_{\pm 55.91}$                                                    & $323.45_{\pm 5.89}$                                                                             \\
700                  & $578.89_{\pm 19.17}$                                                     & $1339.04_{\pm 40.40}$                                                    & $320.95_{\pm 14.18}$                                                                             \\
999                  & $557.78_{\pm 1.92}$                                                      & $1345.81_{\pm 36.31}$                                                    & $321.90_{\pm 10.19}$                                                                             \\ \bottomrule
\end{tabular}
}
\caption{An ablation study of total noise steps $T$ on the MME benchmark.}
\label{tab:ablation1}
\end{table}

%% file: tables/ablation2.tex
\begin{table}[t]
\centering
\resizebox{1\linewidth}{!}{%
\begin{tabular}{@{}c|ccc@{}}
\toprule
\textbf{$\alpha$} & \textit{\begin{tabular}[c]{@{}c@{}}Hallucination Subset\\ Total Scores\end{tabular}} & \textit{\begin{tabular}[c]{@{}c@{}}Perception Subset\\ Total Scores\end{tabular}} & \textit{\begin{tabular}[c]{@{}c@{}}Recognition Subset\\ Total Scores\end{tabular}} \\ \midrule
0.25              & $583.89_{\pm 19.32}$ & $1322.25_{\pm 32.58}$ & $330.24_{\pm 13.60}$ \\
0.5               & $580.56_{\pm 17.11}$ & $1315.49_{\pm 27.28}$ & $333.45_{\pm 5.77}$ \\
0.75              & $578.33_{\pm 29.49}$ & $1312.93_{\pm 39.31}$ & $330.95_{\pm 13.58}$ \\
1.0               & $591.67_{\pm 36.06}$ & $1340.89_{\pm 55.91}$ & $323.45_{\pm 5.89}$ \\ \bottomrule
\end{tabular}
}
\caption{An ablation study of $\alpha$ on the MME benchmark.}
\label{tab:ablation2}
\end{table}

%% file: tables/ablation3.tex
\begin{table}[t]
\centering
\resizebox{1\linewidth}{!}{%
\begin{tabular}{@{}c|ccc@{}}
\toprule
\textbf{$\beta$} & \textit{\begin{tabular}[c]{@{}c@{}}Hallucination Subset\\ Total Scores\end{tabular}} & \textit{\begin{tabular}[c]{@{}c@{}}Perception Subset\\ Total Scores\end{tabular}} & \textit{\begin{tabular}[c]{@{}c@{}}Recognition Subset\\ Total Scores\end{tabular}} \\ \midrule
0                 & $577.22_{\pm 11.10}$                                                                  & $1299.04_{\pm 39.30}$                                                              & $302.98_{\pm 19.82}$                                                                  \\
0.001             & $574.44_{\pm 6.31}$                                                                  & $1298.71_{\pm 40.24}$                                                             & $289.88_{\pm 14.02}$                                                               \\
0.01              & $583.33_{\pm 18.78}$                                                                 & $1324.44_{\pm 37.84}$                                                             & $327.38_{\pm 17.11}$                                                               \\
0.1               & $591.67_{\pm 36.06}$                                                                 & $1340.89_{\pm 55.91}$                                                             & $323.45_{\pm 5.89}$                                                                \\
0.2               & $591.67_{\pm 7.26}$                                                                  & $1343.06_{\pm 13.06}$                                                             & $328.57_{\pm 16.37}$                                                               \\
0.5               & $635.00_{\pm 7.64}$                                                                  & $1474.02_{\pm 15.53}$                                                             & $331.43_{\pm 13.03}$                                                               \\
% 1                 & $648.33_{\pm 0.00}$                                                                  & $1508.98_{\pm 0.25}$                                                              & $359.64_{\pm 3.09}$                                                                \\ 
\bottomrule
\end{tabular}
}
\caption{Ablation studies of $\beta$ on the MME benchmark.}
\label{tab:ablation3}
\end{table}

%% file: tables/ablation4.tex
\begin{table*}[tp]
\centering
\begin{tabular}{lcccc|c}
\toprule
\multicolumn{1}{l}{Sampling Strategy }          &  w. VCD    & Accuracy & Precision & Recall & F1 Score    \\ \midrule
\multirow{2}{*}{Top P}       & No  & $84.91_{\pm 0.25}$     & $94.73_{\pm 0.30
}$     & $73.93_{\pm 0.52}$  & $83.05_{\pm 0.32}$ \\
                              & Yes & $\textbf{87.82}_{\pm 0.66}$    & $91.17_{\pm 0.57}$     & $83.76_{\pm 0.87}$  & $\textbf{87.31}_{\pm 0.72}$  \\ \midrule
\multirow{2}{*}{Top K}         & No  & $83.04_{\pm 0.16}$    & $91.84_{\pm 0.15}$     & $72.53_{\pm 0.44}$  & $81.05_{\pm 0.24}$ \\
                              & Yes & $\textbf{87.49}_{\pm 0.56}$    & $91.09_{\pm 0.53}$     & $83.11_{\pm 0.71}$  & $\textbf{86.92}_{\pm 0.60}$ \\ \midrule
\multirow{2}{*}{Greedy}       & No  & $87.10_{\pm 0.00}$    & $97.33_{\pm 0.00}$     & $76.29_{\pm 0.00}$   & $85.54_{\pm 0.00}$  \\
                              & Yes & $\textbf{88.49}_{\pm 0.28}$    & $91.78_{\pm 0.28}$     & $84.56_{\pm 0.44}$  & $\textbf{88.02}_{\pm 0.30}$ \\ \midrule
\multirow{2}{*}{\begin{tabular}[c]{@{}l@{}}Top K+Temperature 0.7\end{tabular}} & No  & $85.17_{\pm 0.12}$     & $94.82_{\pm 0.12}$     & $74.40_{\pm 0.35}$  & $83.38_{\pm 0.17}$ \\
                              & Yes & $\textbf{87.94}_{\pm 0.51}$    & $91.21_{\pm 0.49}$     & $83.98_{\pm 0.60}$  & $\textbf{87.45}_{\pm 0.54}$ \\ \midrule
\multirow{2}{*}{\begin{tabular}[c]{@{}l@{}}Top K+Temperature 1.5\end{tabular}} & No  & $79.28_{\pm 0.22}$    & $86.48_{\pm 1.12}$     & $69.42_{\pm 0.91}$  & $77.01_{\pm 0.22}$ \\
                              & Yes & $\textbf{86.97}_{\pm 0.50}$    & $90.96_{\pm 0.64}$     & $82.09_{\pm 0.41}$  & $\textbf{86.30}_{\pm 0.51}$ \\
\bottomrule
\end{tabular}
\caption{An ablation study of different sampling strategies.}
\label{tab:ablation4}
\end{table*}

%% file: tables/pope_upscale.tex
\begin{table*}[]
\centering
\resizebox{0.8\linewidth}{!}{%
\begin{tabular}{@{}cllllll|l@{}}
\toprule
\textbf{Dataset}         & \textbf{POPE}                         & \textbf{Model}                     & \textbf{Decoding} & Accuracy  & Precision & Recall    & F1 Score \\ \midrule
\multirow{12}{*}{MSCOCO} & \multirow{4}{*}{\textit{Random}}      & \multirow{2}{*}{LLaVA1.5(13B)}     & Regular           &$83.31_{\pm 0.32}$ &$91.46_{\pm 0.38}$ &$73.48_{\pm 0.75}$ &$81.49_{\pm 0.43}$ \\
                         &                                       &                                    & VCD               &$\textbf{87.39}_{\pm 0.32}$ &$92.68_{\pm 0.36}$ &$81.19_{\pm 0.63}$ &$\textbf{86.55}_{\pm 0.41}$ \\
                         &                                       & \multirow{2}{*}{InstructBLIP(13B)} & Regular           &$82.36_{\pm 0.59}$ &$86.93_{\pm 0.85}$ &$76.19_{\pm 1.05}$ &$81.20_{\pm 0.68}$ \\
                         &                                       &                                    & VCD               &$\textbf{84.53}_{\pm 0.38}$ &$88.55_{\pm 0.54}$ &$79.32_{\pm 0.44}$ &$\textbf{83.68}_{\pm 0.40}$ \\ \cmidrule(l){2-8} 
                         & \multirow{4}{*}{\textit{Popular}}     & \multirow{2}{*}{LLaVA1.5(13B)}     & Regular           &$82.47_{\pm 0.55}$ &$89.55_{\pm 0.92}$ &$73.53_{\pm 0.78}$ &$80.75_{\pm 0.61}$ \\
                         &                                       &                                    & VCD               &$\textbf{85.74}_{\pm 0.25}$ &$89.33_{\pm 0.52}$ &$81.19_{\pm 0.63}$ &$\textbf{85.06}_{\pm 0.29}$ \\
                         &                                       & \multirow{2}{*}{InstructBLIP(13B)} & Regular           &$79.07_{\pm 0.66}$ &$81.11_{\pm 0.70}$ &$75.79_{\pm 1.27}$ &$78.35_{\pm 0.78}$ \\
                         &                                       &                                    & VCD               &$\textbf{81.47}_{\pm 0.42}$ &$82.89_{\pm 0.64}$ &$79.32_{\pm 0.44}$ &$\textbf{81.07}_{\pm 0.39}$ \\ \cmidrule(l){2-8} 
                         & \multirow{4}{*}{\textit{Adversarial}} & \multirow{2}{*}{LLaVA1.5(13B)}     & Regular           &$80.00_{\pm 0.52}$ &$84.46_{\pm 0.73}$ &$73.53_{\pm 0.76}$ &$78.62_{\pm 0.58}$ \\
                         &                                       &                                    & VCD               &$\textbf{81.92}_{\pm 0.44}$ &$82.40_{\pm 0.42}$ &$81.17_{\pm 0.65}$ &$\textbf{81.78}_{\pm 0.47}$ \\
                         &                                       & \multirow{2}{*}{InstructBLIP(13B)} & Regular           &$76.57_{\pm 0.75}$ &$77.00_{\pm 0.83}$ &$75.79_{\pm 0.80}$ &$76.39_{\pm 0.75}$ \\
                         &                                       &                                    & VCD               &$\textbf{79.56}_{\pm 0.41}$ &$79.67_{\pm 0.59}$ &$79.39_{\pm 0.50}$ &$\textbf{79.52}_{\pm 0.38}$ \\ \bottomrule
\end{tabular}
}
\caption{Results of 13B-sized LLaVA1.5 and InstructBLIP variants on the POPE metric. The best performance of each setting is \textbf{bolded}. }
\label{tab: pope_upscale}
\end{table*}

%% file: tables/mme_percep.tex
\begin{table*}[!ht]
\centering
\resizebox{1\linewidth}{!}{%
\begin{tabular}{@{}llllllllllll|l@{}}
\toprule
Model                         & Decoding & \multicolumn{1}{c}{\textit{Existence}} & \multicolumn{1}{c}{\textit{Count}} & \multicolumn{1}{c}{\textit{Position}} & \multicolumn{1}{c}{\textit{Color}} & \multicolumn{1}{c}{\textit{Posters}} & \multicolumn{1}{c}{\textit{Celebrity}} & \multicolumn{1}{c}{Scene} & \multicolumn{1}{c}{Landmark} & \multicolumn{1}{c}{Artwork} & \multicolumn{1}{c|}{OCR} & \multicolumn{1}{c}{\textit{\textbf{\begin{tabular}[c]{@{}c@{}}Percetion \\ Total\end{tabular}}}} \\ \midrule
\multirow{2}{*}{LLaVA1.5}     & Regular  & $175.67_{\pm 7.51}$ & $124.67_{\pm 19.59}$ & $114.00_{\pm 9.32}$ & $151.00_{\pm 10.45}$ & $127.82_{\pm 7.13} $& $113.59_{\pm 3.43}$ & $148.30_{\pm 3.49}$ & $129.95_{\pm 5.33}$ & $102.20_{\pm 4.70}$ & $92.00_{\pm 31.29}$ & $1279.19_{\pm 37.09}$ \\
                              & VCD      & $\textbf{184.66}_{\pm 6.81}$ & $\textbf{138.33}_{\pm 15.68}$ & $\textbf{128.67}_{\pm 7.21}$ & $\textbf{153.00}_{\pm 7.58}$ & $\textbf{132.11}_{\pm 6.53}$ & $\textbf{120.94}_{\pm 7.57}$ & $\textbf{152.20}_{\pm 0.21}$ & $\textbf{140.45}_{\pm 6.73}$ & $\textbf{109.60}_{\pm 2.66}$ & $\textbf{104.00}_{\pm 30.96}$ & $\textbf{1363.96}_{\pm 40.58}$ \\ \midrule
\multirow{2}{*}{Qwen-VL}      & Regular  & $155.00_{\pm 3.54}$ & $127.67_{\pm 13.36}$ & $\textbf{131.67}_{\pm 7.73}$ & $173.00_{\pm 9.75}$ & $137.76_{\pm 2.49}$ & $116.24_{\pm 2.58}$ & $\textbf{150.17}_{\pm 2.80}$ & $158.00_{\pm 2.40}$ & $\textbf{125.75}_{\pm 5.74}$ & $\textbf{89.50}_{\pm 7.37}$ & $1364.74_{\pm 30.78}$ \\
                              & VCD      & $\textbf{156.00}_{\pm 6.52}$ & $\textbf{131.00}_{\pm 6.19}$ & $128.00_{\pm 3.61}$ & $\textbf{181.67}_{\pm 5.14}$ & $\textbf{142.45}_{\pm 2.96}$ & $\textbf{137.35}_{\pm 2.45}$ & $149.10_{\pm 2.51}$ & $\textbf{163.95}_{\pm 1.77}$ & $127.65_{\pm 2.81}$ & $86.00_{\pm 3.35}$ & $\textbf{1403.17}_{\pm 14.57}$ \\ \midrule
\multirow{2}{*}{InstructBLIP} & Regular  & $141.00_{\pm 13.97}$ & $75.33_{\pm 14.16}$ & $\textbf{66.67}_{\pm 3.91}$ & $97.33_{\pm 16.94}$ & $109.66_{\pm 6.21}$ & $87.50_{\pm 6.80}$ & $128.74_{\pm 3.13}$ & $100.55_{\pm 3.33}$ & $94.10_{\pm 5.05}$ & $83.50_{\pm 19.25}$ & $1223.72_{\pm 86.59}$ \\
                              & VCD      & $\textbf{168.33}_{\pm 11.55}$ & $\textbf{92.33}_{\pm 8.47}$ & $64.00_{\pm 6.73}$ & $\textbf{123.00}_{\pm 11.27}$ & $\textbf{121.09}_{\pm 5.12}$ & $\textbf{118.71}_{\pm 3.93}$ & $\textbf{149.65}_{\pm 1.46}$ & $\textbf{123.65}_{\pm 1.89}$ & $\textbf{110.60}_{\pm 2.89}$ & $\textbf{96.50}_{\pm 8.94}$ & $\textbf{1447.19}_{\pm 25.43}$ \\ \bottomrule
\end{tabular}
}
\caption{Results on all MME perception-related tasks. The best performance of each setting is \textbf{bolded}.}
\label{tab:mme_perception}
%\hspace{-8cm}
\end{table*}

%% file: tables/mme_recog.tex
\begin{table*}[ht]
\centering
\begin{tabular}{@{}llllll|l@{}}
\toprule
Model                         & Decoding & \multicolumn{1}{c}{\textit{\begin{tabular}[c]{@{}c@{}}Common Sense\\ Reasoning\end{tabular}}} & \multicolumn{1}{c}{\textit{\begin{tabular}[c]{@{}c@{}}Numerical\\ Calculation\end{tabular}}} & \multicolumn{1}{c}{\textit{\begin{tabular}[c]{@{}c@{}}Text\\ Translation\end{tabular}}} & \multicolumn{1}{c|}{\textit{\begin{tabular}[c]{@{}c@{}}Code\\ Reasoning\end{tabular}}} & \multicolumn{1}{c}{\textit{\textbf{\begin{tabular}[c]{@{}c@{}}Recognition\\ Total\end{tabular}}}} \\ \midrule
\multirow{2}{*}{LLaVA1.5}     & Regular  & $106.43_{\pm 9.04}$ & $\textbf{72.50}_{\pm 15.51}$ & $\textbf{95.50}_{\pm 12.80}$ & $78.50_{\pm 22.12}$ & $352.93_{\pm 27.98}$ \\
                              & VCD      & $\textbf{111.29}_{\pm 7.06}$ & $68.50_{\pm 16.64}$ & $89.50_{\pm 5.97}$ & $\textbf{84.00}_{\pm 25.35}$ & $\textbf{353.29}_{\pm 36.19}$ \\ \midrule
\multirow{2}{*}{Qwen-VL}      & Regular  & $109.86_{\pm 10.31}$ & $\textbf{60.00}_{\pm 6.37}$ & $83.00_{\pm 11.91}$ & $\textbf{67.50}_{\pm 10.16}$ & $\textbf{320.36}_{\pm 26.00}$ \\
                              & VCD      & $\textbf{114.39}_{\pm 5.83}$ & $54.00_{\pm 9.62}$ & $\textbf{85.00}_{\pm 7.29}$ & $64.50_{\pm 7.37}$ & $317.89_{\pm 11.59}$ \\ \midrule
\multirow{2}{*}{InstructBLIP} & Regular  & $79.57_{\pm 6.81}$ & $62.86_{\pm 11.23}$ & $55.00_{\pm 10.75}$ & $70.00_{\pm 10.75}$ & $267.43_{\pm 10.42}$ \\
                              & VCD      & $\textbf{109.71}_{\pm 7.31}$ & $\textbf{66.00}_{\pm 16.45}$ & $\textbf{69.00}_{\pm 11.54}$ & $\textbf{74.50}_{\pm 20.26}$ & $\textbf{319.21}_{\pm 20.60}$ \\ \bottomrule
\end{tabular}
\caption{Results on all MME recognition-related tasks. The best performance of each setting is \textbf{bolded}.}
\label{tab:mme_recognition}
\end{table*}

%% file: figs/case_hallu.tex
\begin{figure*}[tp]
    \centering
    \includegraphics[width=1\linewidth]{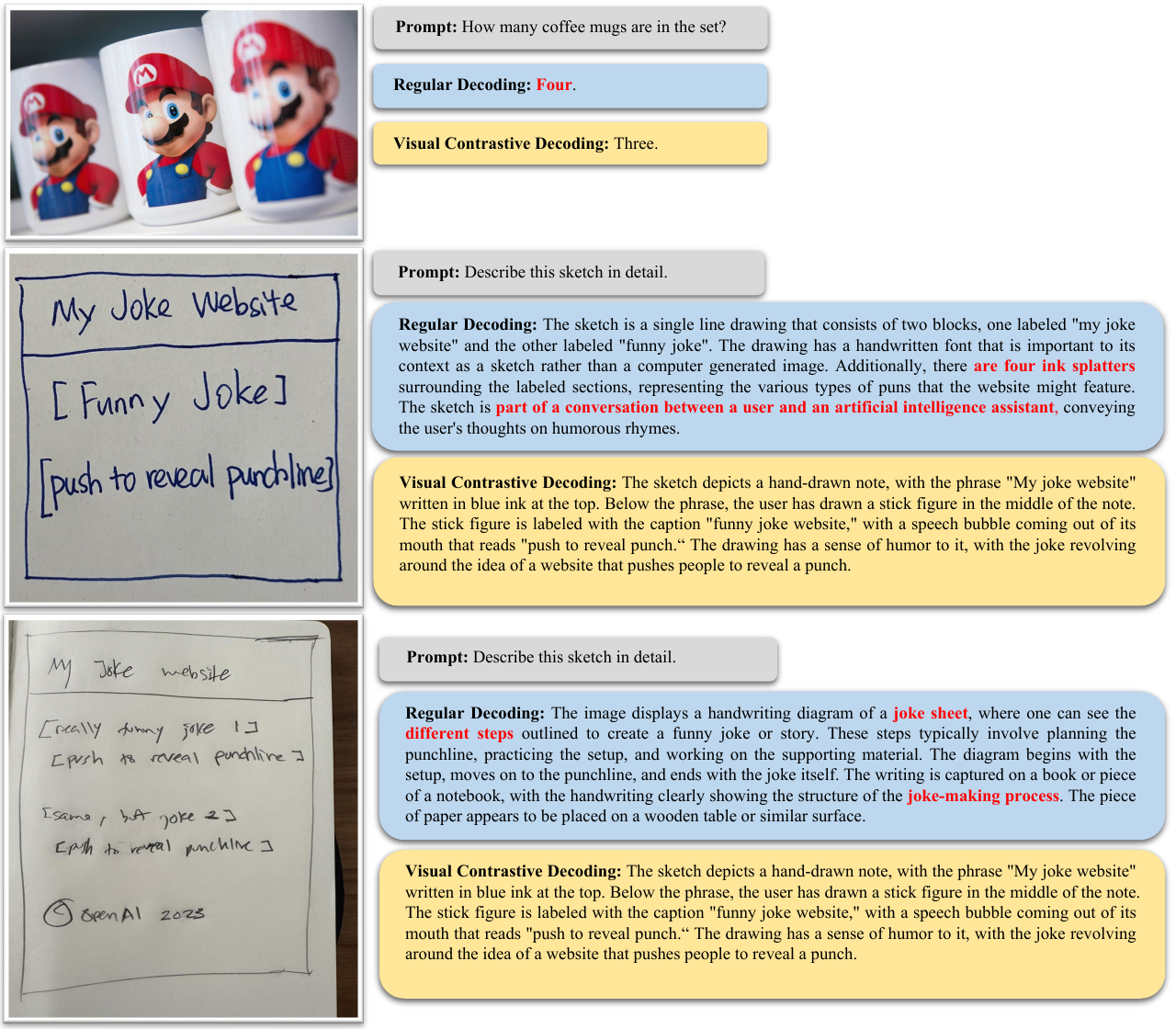}
    \caption{More examples from LLaVA-Bench of our proposed VCD for hallucination corrections. Hallucinated objects from LVLM's regular decoding are highlighted in \color{red}{red}.}
    \label{fig:case_hallu}
    \vspace{-0.4cm}
\end{figure*}

%% file: figs/case_general.tex
\begin{figure*}[tp]
    \centering
    \includegraphics[width=1\linewidth]{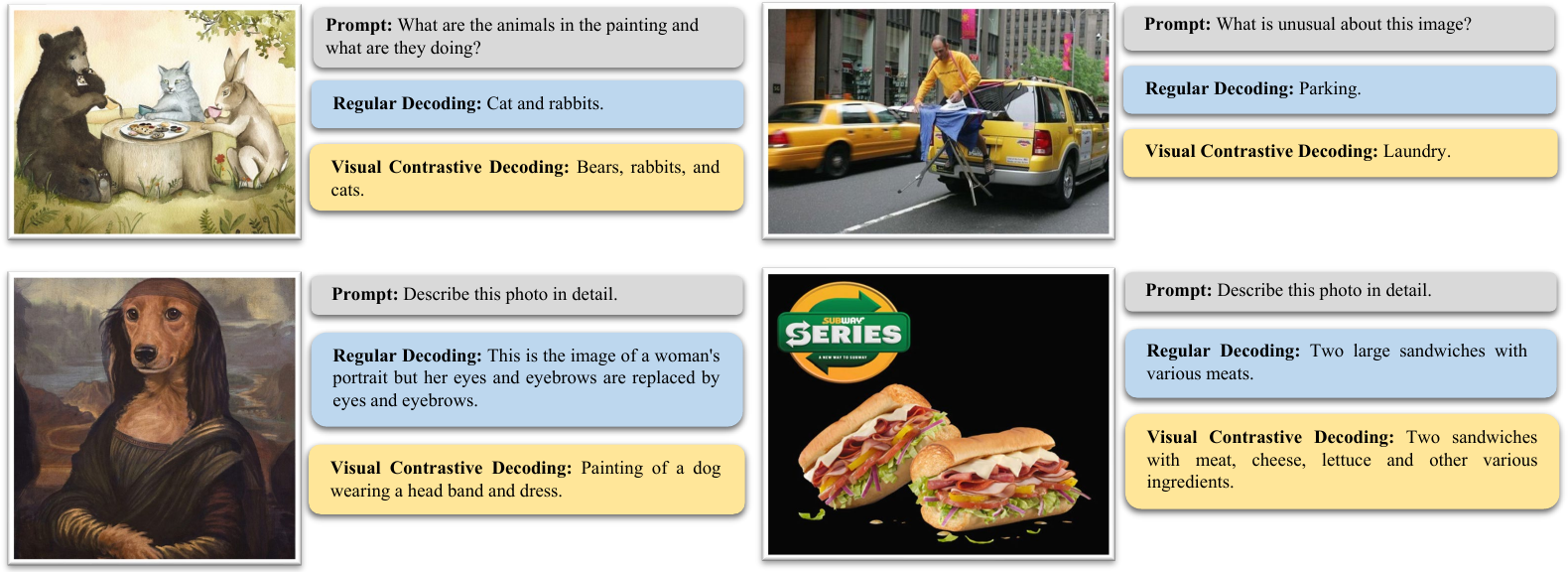}
    \caption{More examples from LLaVA-Bench of our proposed VCD for enhanced general perception and recognition capacities.}
    \label{fig:case_general}
    \vspace{-0.4cm}
\end{figure*}

%% file: tables/GPT4_prompt.tex
\begin{table*}[h!]\centering
\begin{minipage}{0.95\textwidth}
%\vspace{0mm}    
\centering
\begin{tcolorbox} 
    \centering
   
     %\hspace{-4mm}
      \small
    \begin{tabular}{p{0.95\textwidth}} \hline \\
   \textbf{Description:} \\    
   
   AI that scores image description accuracy and detailedness.

   \\ \midrule

   \textbf{Instructions:} \\   
   
You are an AI designed to evaluate and score the performance of two AI assistants in describing a given image. Your primary focus is on the accuracy and detailedness of their descriptions. You will assess the accuracy by checking for hallucinations - any part of the description that is inconsistent with the image content. For detailedness, you will consider how rich the response is in necessary details, excluding any hallucinated parts. You will provide scores on a scale from 1 to 10 for each assistant separately, based on these criteria. After scoring, you will offer an explanation for your evaluation, ensuring it is free from bias and not influenced by the order of presentation of the responses.
\\ \\
Input format: \\ \\
\lbrack{}Assistant 1\rbrack{}\\
 \{Response 1\}  \\
\lbrack{}End of Assistant 1\rbrack{} \\
\\
\lbrack{}Assistant 2\rbrack{} \\
 \{Response 2\}\\
\lbrack{}End of Assistant 2\rbrack{} \\
\\
Output format:\\
\\
Accuracy:\\
Scores of the two answers:\\
Reason:\\
\\
Detailedness:\\
Scores of the two answers:\\
Reason:\\ \\

\bottomrule
    \end{tabular}
\end{tcolorbox}
%\vspace{-2mm}
\caption{The configuration to build an image-description evaluator with GPT-4V}
    \label{tab:prompt_evaluation}
\end{minipage}
\end{table*}

%% file: figs/gpt4_evaluator_case.tex
\begin{figure*}[tp]
    \centering
    \includegraphics[width=0.95\linewidth]{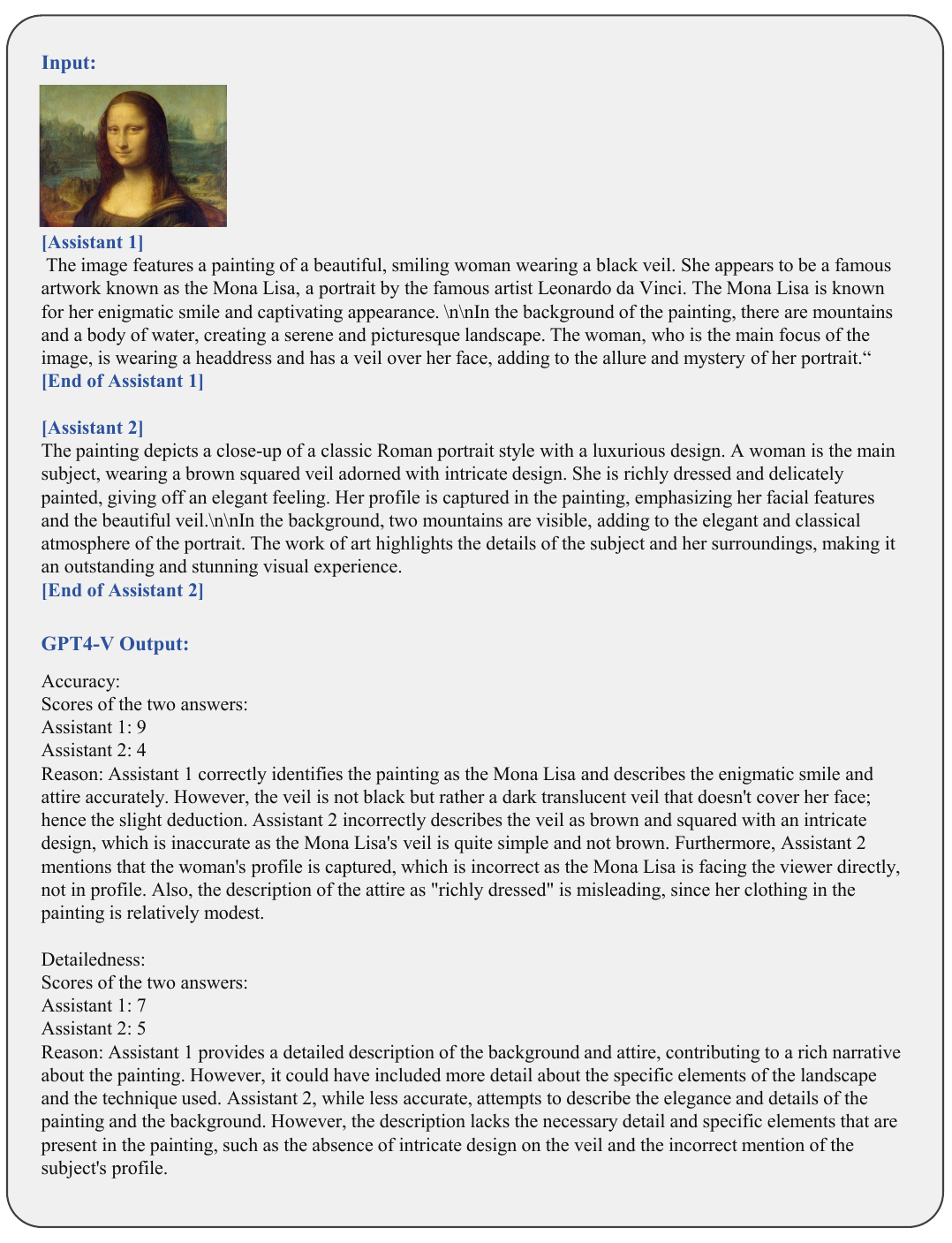}
    \caption{Case illustrating the evaluation of GPT-4V in open-ended generation task. ``Assistant 1'' and ``Assistant 2'' correspond to
``visual contrastive decoding'' and ``regular decoding''.}
    \label{fig: gpt4_evaluator_case}
    % \vspace{-0.4cm}
\end{figure*}